\documentclass[11pt]{article}
\usepackage{acl}
\usepackage{times}
\usepackage{latexsym}
\usepackage[T1]{fontenc}
\usepackage[utf8]{inputenc}
\usepackage{microtype}
\usepackage{inconsolata}
\usepackage{graphicx}
\usepackage{amsmath,amssymb}
\usepackage{float}
\usepackage{multirow}
\usepackage{makecell}
\usepackage{booktabs}
\usepackage{rotating}
\usepackage{array}
\usepackage{tabularx}
\usepackage{longtable}
\usepackage{supertabular}
\usepackage{hyperref}
\usepackage{enumitem}
\usepackage{xcolor}
\definecolor{darkblue}{RGB}{0,0,120}

\usepackage{tcolorbox}
\hypersetup{
    colorlinks=true,
    urlcolor=black,
    linkcolor=darkblue,
    citecolor=darkblue
}

\title{Perceive Before Reasoning: A Pre-Reasoning Perception Framework for Efficient and Reliable Proactive Mobile Agents}

  \author{
  {\bfseries Zhijie Ding$^{1,2}$\thanks{These authors contributed equally.}\;
  Weinan Hong$^{1,3}$\footnotemark[1]\;
  Zicheng Zhu$^{1,4}$\footnotemark[1]\;
  Lei Li$^{1}$\footnotemark[1]\;
  Dezhi Kong$^{1}$} \\
  {\bfseries Hao Wang$^{1}$\;
  Peng Zhou$^{1}$\;
  Xuchu Jiang$^{2}$\thanks{Corresponding authors.}\;
  Jiaming Xu$^{1}$\footnotemark[2]} \\
  \normalfont
  dingzhijie@stu.zuel.edu.cn
  \quad
  hongwn24@mails.jlu.edu.cn
  \quad
  xujiaming1@xiaomi.com\\
  \normalfont
  $^{1}$HyperAI Team, Xiaomi Corporation \quad
  $^{2}$Zhongnan University of Economics and Law \\
  \normalfont
  $^{3}$Jilin University \quad
  $^{4}$The Chinese University of Hong Kong, Shenzhen
  }

\begin{document}
\maketitle

\begin{abstract}

Multimodal large language models (MLLMs) have substantially advanced mobile agents, yet proactive mobile assistance remains challenging because agents must decide \emph{when} to intervene before determining \emph{how} to assist. Existing systems often implement these two decisions within a unified MLLM-based pipeline, leading to goal misalignment between conservative intervention filtering and comprehensive assistance generation, as well as redundant inference when the agent should remain silent. To address these limitations, we propose the \textbf{Pre-Reasoning Perception Framework (PRPF)}, a two-stage framework built on perceiving before reasoning. PRPF introduces a lightweight Multimodal Proactive Perceptor (MPP) for intervention gating and context compression, and activates the Proactive Agent Reasoner (PAR) only when intervention is warranted. Experiments on the ProactiveMobile benchmark show that PRPF substantially reduces false trigger rates (FTR) while improving success rates (SR) and inference efficiency over the ProactiveMobile baseline. 

\end{abstract}
\section{Introduction}

Multimodal large language models (MLLMs), particularly vision-language models (VLMs), have substantially advanced mobile agents by enabling them to perceive mobile interfaces and execute user instructions~\citep{qwen35blog, hurst2024gpt, zhang2025appagent}. However, mobile agents are increasingly expected to move beyond reactive instruction following toward proactive assistance, where they anticipate user needs and intervene without explicit prompts~\citep{proactiveagent, proactivemobile}. Existing studies commonly formulate proactive assistance as a \emph{when--how} problem, in which an agent must first determine \emph{when} to intervene and then decide \emph{how} to assist~\citep{yang2026contextagent,xie2026pask,vigil}.

\begin{figure}[t]
    \centering
    \includegraphics[width=\columnwidth]{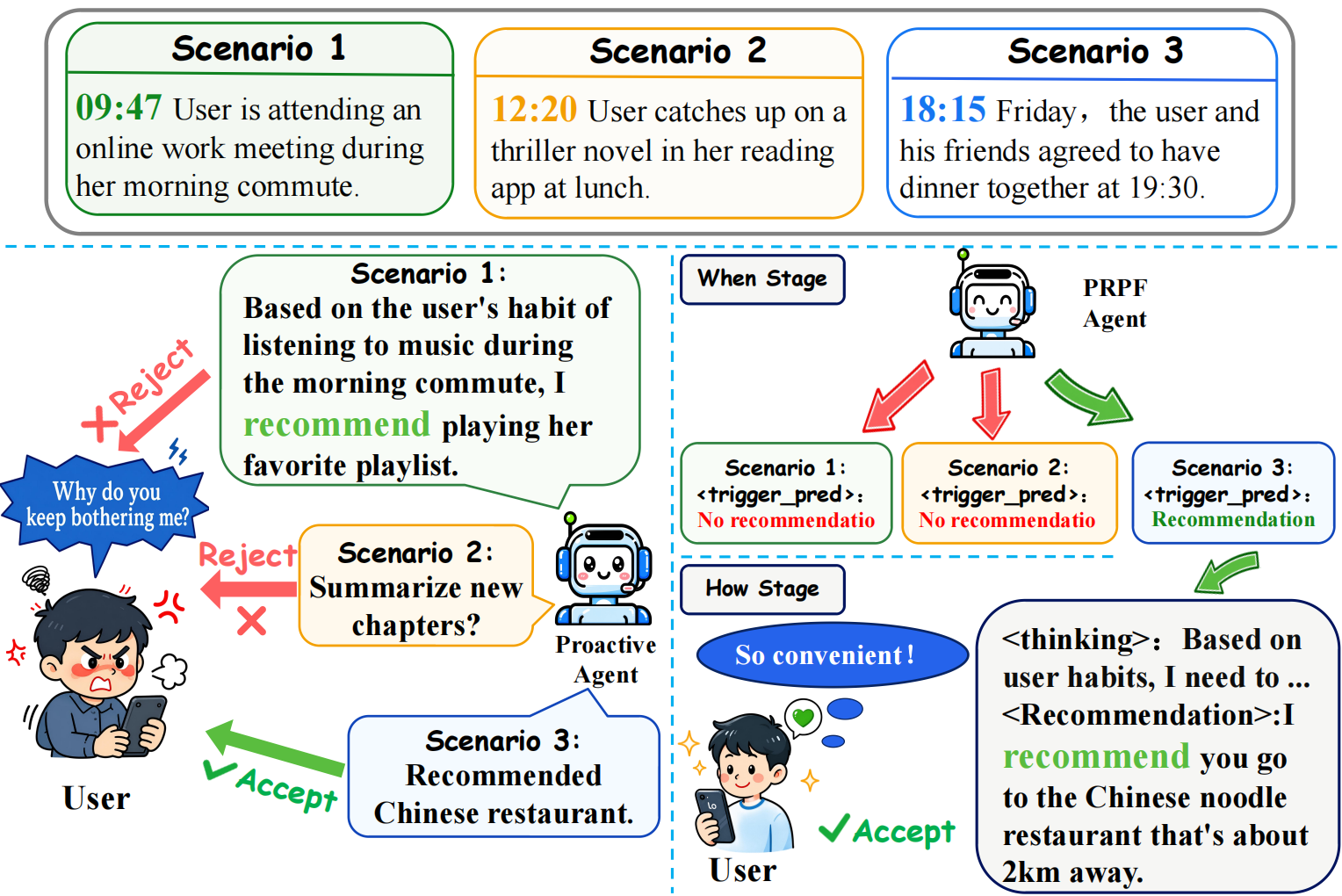}
    \caption{Comparison between unified proactive reasoning and PRPF. Left: Many existing proactive agents couple \emph{when} and \emph{how} within a unified framework. Right: PRPF decouples these two tasks and filters out cases that do not require recommendation before reasoning, thereby reducing unnecessary interruptions.}
    \label{fig:comparison}
\end{figure}

Recent systems typically realize this formulation within a unified VLM-based agent. For example, ProAgentBench treats when to assist and how to assist as timing judgment and content generation subtasks handled by the same fine-tuned backbone~\citep{tang2026proagentbench}. Similarly, PARE-Bench frames proactive assistance as an Observe--Execute process, where observation and execution are realized by switching tool-use modes within one VLM-based assistant~\citep{PARE}. Despite differences in implementation, these systems generally couple the \emph{when} and \emph{how} decisions within a unified proactive-agent framework. As illustrated in Figure~\ref{fig:comparison}, existing proactive agents invoke the same reasoning pipeline for deciding whether to intervene and determining how to assist, whereas PRPF decouples these two decisions by filtering cases that do not require recommendation before the reasoning stage. This unified design gives rise to two key limitations:

\textbf{Goal misalignment}. The when decision requires conservative and highly discriminative judgment to suppress false triggers under asymmetric intervention costs, whereas the \emph{how} decision requires broad multimodal reasoning and flexible content generation. Coupling these goals within a single VLM-agent makes it difficult to simultaneously achieve conservative intervention filtering and comprehensive assistance generation. As observed in Proactive Agent and PRISM, even strong models struggle to maintain the low false-trigger rates required for satisfactory mobile user experience~\citep{proactiveagent, fu2026prism}. Excessive interventions disrupt user workflows, while overly conservative policies risk reducing proactive agents to passive assistants.

\textbf{Inference inefficiency}. Existing single-stage VLM agents execute the full reasoning pipeline regardless of whether intervention is ultimately needed. As a result, large models are activated even when the correct behavior is to remain silent, incurring unnecessary long-context and multi-turn reasoning costs~\citep{Gao2024CostEfficientLL, yang2026proagentharnessingondemandsensory}. This inefficiency becomes more severe in mobile environments with diverse scenarios and large function spaces. For example, ProactiveMobile requires reasoning over schemas spanning 14 scenarios, substantially increasing prompt length and inference cost~\citep{proactivemobile}. ContextAgent similarly reports higher computational costs with limited performance gains under complex contextual settings~\citep{yang2026contextagent}.

To address these limitations, we propose the \textbf{P}re-\textbf{R}easoning \textbf{P}erception \textbf{F}ramework (\textbf{PRPF}), a mobile proactive intelligence framework built on the principle of perceiving before reasoning. Rather than using a large VLM to jointly determine when to intervene and how to assist, PRPF separates the two stages at the architectural level. A lightweight perceptual front-end first performs intervention gating and candidate function compression, assigning the conservative when judgment to an efficient discriminative module. The VLM-based reasoner is then activated only when intervention is warranted, allowing it to focus on the generative \emph{how}-stage. This separation aligns each module with its corresponding goal while avoiding full VLM inference in cases where the agent should remain silent.

Concretely, PRPF introduces a \textbf{Multimodal Proactive Perceptor (MPP)}, a lightweight multimodal fusion encoder with short- and long-term pathways that operates before large-model reasoning. MPP first decides whether the agent should intervene or remain silent, thereby reducing false triggers at the source. It then selects the top-$K$ intent scenarios and takes the union of their
associated functions to form a compressed function pool, filtering irrelevant long-tail contexts and reducing the reasoning burden for the subsequent stage. Given this compact evidence, the \textbf{Proactive Agent Reasoner (PAR)} conducts focused deep reasoning to generate the final proactive recommendation. As a result, MPP serves as a pre-reasoning perceptual cutoff, while PAR concentrates its reasoning capacity on cases where proactive assistance is actually needed.  Experiments on the ProactiveMobile benchmark show that PRPF improves the success rate from 20.82\% to 41.15\%, reduces the false trigger rate from 13.76\% to 7.21\%, and cuts expected inference compute by 69.3\%. These results show that PRPF improves proactive recommendation and tool invocation accuracy while reducing false interruptions by better identifying when to remain silent. In summary, our contributions are as follows:

\begin{itemize}
\item 
We propose PRPF, a two-stage pre-reasoning perception framework that architecturally decouples \emph{when} judgment from \emph{how}-stage reasoning for proactive mobile agents.

\item 
We design MPP, a lightweight and plug-and-play multimodal perceptor for intervention gating and context compression, together with PAR, a Proactive Agent Reasoner for complete and focused \emph{how}-stage reasoning.

\item 
Extensive experiments on the ProactiveMobile benchmark demonstrate that PRPF substantially reduces false trigger rates while improving success rates and inference efficiency.
\end{itemize}

\begin{figure*}[t]
    \centering
    \includegraphics[width=\textwidth]{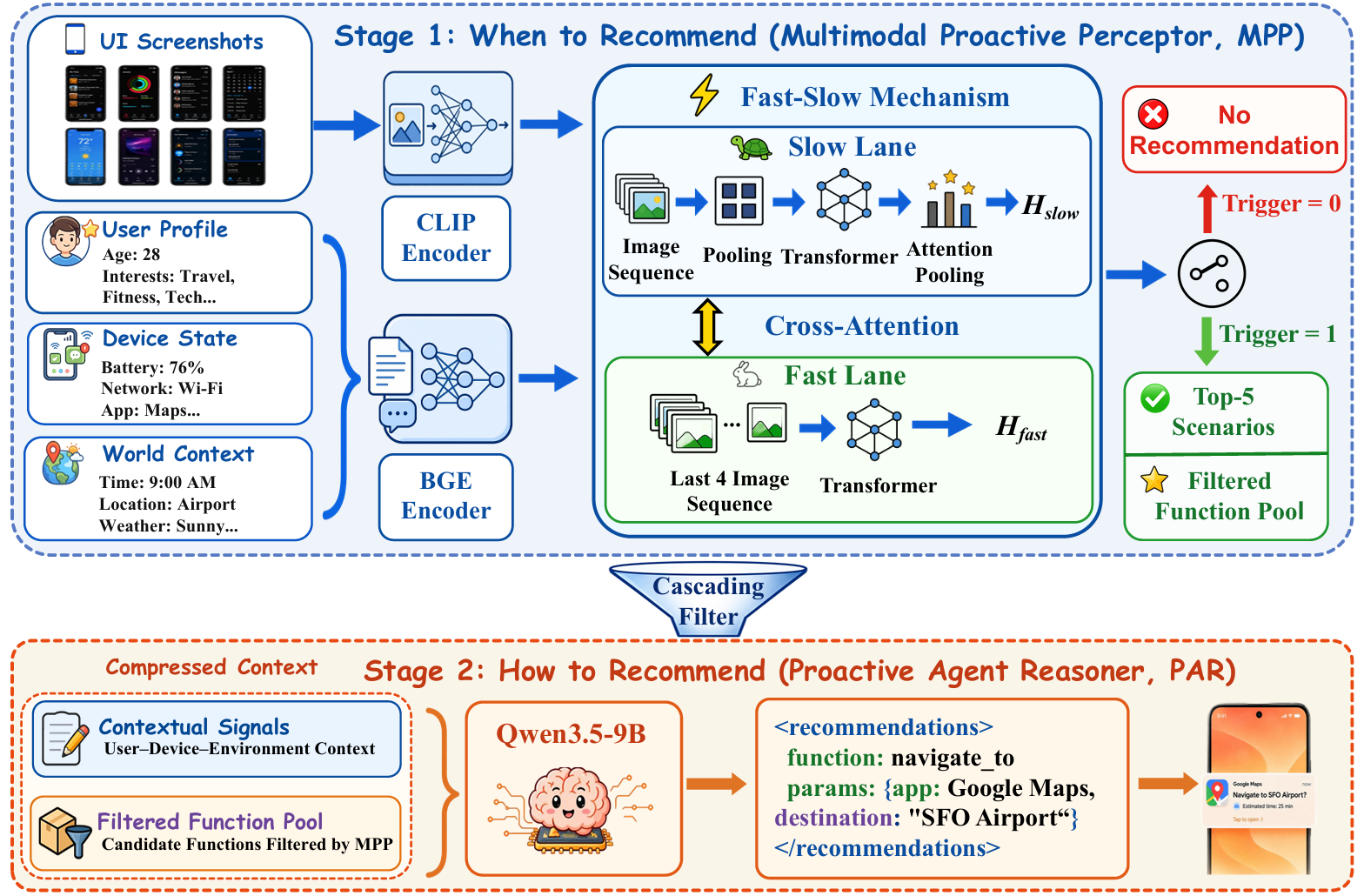}
    \caption{Overall architecture of PRPF. MPP determines whether proactive intervention is necessary, selects top-$K$ intent scenarios to construct the compressed function pool, while PAR infers user intent and generates executable recommendations from the pool.}
    \label{fig:framework}
\end{figure*}

\section{Related Work}
\label{sec:related_work}

\subsection{Proactive Agents}
\label{sec:related_proactive_agents}

Proactive-agent research extends LLM agents from instruction following to active assistance, where agents must judge intervention timing and generate useful assistance from ongoing context~\citep{proactiveagent,tang2026proagentbench,fu2026prism}. Prior work studies this distinction through intervention-timing prediction, assistance-content generation, cost-sensitive selective intervention, and staged proactive behavior~\citep{tang2026proagentbench,fu2026prism,PARE,xie2026pask}. Proactive GUI and mobile settings make this problem concrete because user intent must be inferred from interface trajectories before an explicit command is issued. Existing benchmarks study intent recommendation over GUI trajectories and executable function-sequence generation from on-device context~\citep{chai2026pira,proactivemobile}. Using ProactiveMobile as the evaluation setting, PRPF focuses on the architectural separation between lightweight pre-reasoning intervention perception and heavy VLM-based assistance reasoning.

\subsection{GUI Perception and Efficient Reasoning}
\label{sec:related_gui_reasoning}

Mobile and GUI agents provide the perception and execution substrate for proactive assistance, but most existing systems remain reactive. Prior work has advanced smartphone operation, mobile task execution, screenshot-based GUI understanding, and visual grounding under explicit instructions~\citep{zhang2025appagent,wang2024mobile,rawles2025androidworld,deng2024mobile,qin2025uitarspioneeringautomatedgui,cheng2024seeclick}. These advances improve how an agent perceives and acts on a requested GUI task, but not whether a continuously observed mobile context warrants intervention. Efficiency-oriented work further introduces intermediate perception, adaptive visual processing, region selection, or model routing before expensive reasoning~\citep{wu2025smoothing,mehrotra2025ishift,tang2025think,liu2026drs,ong2024routellm,xu2025learning}. These methods use front-end selection, but their decision signals are typically task difficulty, region relevance, or model-call cost for a given query. They do not by themselves resolve the proactive intervention variable: whether a mobile context warrants intervention and which intent candidates should condition subsequent assistance.

\section{Method}
\label{sec:method}

\subsection{Task Definition}
\label{sec:task_definition}

Proactive intelligence on mobile devices requires models to infer users' latent needs before explicit requests and generate executable function-call sequences~\citep{proactivemobile}. Therefore, this task not only requires accurately determining when to trigger recommendations, but also generating function sequences aligned with user intent. Given a mobile interaction sample, its input includes the user profile ($U$), device state ($D$), world information ($W$), and interaction history ($I$). In the multimodal setting, $I$ is represented as a GUI screenshot sequence, whereas in the text setting, it is represented as trajectory text. The model needs to produce an executable function-call sequence from the function set ($F$):
\begin{equation}
\hat{y} =
\begin{cases}
\varnothing \\
\hat{s} = [(\hat{f}_1,\hat{a}_1), \ldots, (\hat{f}_m,\hat{a}_m)]
\end{cases}
\end{equation}
where $\varnothing$ indicates that proactive recommendation should not be triggered under the current state; $\hat{s}$ denotes the function-call sequence predicted by the model; $\hat{f}_m \in {F}$ denotes the $m$-th predicted function, and $\hat{a}_m$ denotes the corresponding function argument.

\subsection{PRPF Framework}

To address the high false-trigger rate and inference inefficiency in proactive mobile intelligence, we propose the Pre-Reasoning Perception Framework (PRPF), as shown in Figure~\ref{fig:framework}. PRPF consists of two parts: 
1) the lightweight Multimodal Proactive Perceptor (MPP), which quickly determines whether the current mobile interaction state should trigger proactive recommendation and predicts the high-level intent scenarios;
2) the Proactive Agent Reasoner (PAR), which generates function-call sequences consistent with users' intent. MPP filters out samples that do not require recommendation; otherwise, it passes the corresponding function pool to PAR for reasoning.

\subsection{Multimodal Proactive Perceptor}
\label{sec:mpp}

MPP consists of a fast--slow dual-channel interaction module and two task-specific MLP heads for trigger gating and intent-scenario prediction. The latter ranks the 14 high-level intent scenarios, and constructs a compressed function pool from the selected scenarios, while PAR generates executable recommendations from this pool.

Taking multimodal data as an example, the textual information and GUI screenshot sequence are first encoded by the text encoder and image encoder, respectively:
\begin{equation}
\resizebox{0.85\columnwidth}{!}{$
H_{\mathrm{text}} = f_{\mathrm{text}}(\{U,D,W\}),
H_{\mathrm{img}} = \{f_{\mathrm{img}}(I_t)\}_{t=1}^{T},
$}
\label{eq:mpp_encoding}
\end{equation}

\noindent where $f_{\mathrm{text}}$ and $f_{\mathrm{img}}$ denote the text encoder and image encoder, respectively; $I_t$ denotes the $t$-th GUI screenshot; and $T$ denotes the length of the complete GUI screenshot sequence.

\noindent The textual and visual features are then projected into a unified dimensional space through linear mappings:
\begin{equation}
c_{\text{text}} = W_c H_{\text{text}}, \quad c_{\text{img}} = W_v H_{\text{img}},
\label{eq:mpp_projection}
\end{equation}

\noindent where $W_c$ and $W_v$ are the linear projection matrices. 
$c_{\text{text}} \in \mathbb{R}^{3 \times d}$ and $c_{\text{img}} \in \mathbb{R}^{T \times d}$ denote the projected textual context representation and screenshot sequence representation, respectively, and $d$ is the unified projection dimension.

The function-call sequence depends on both short-term GUI dynamics and long-term behavioral preferences reflected in the textual context. Accordingly, we construct fast and slow channels to capture these complementary signals.

The slow channel captures long-range behavioral patterns that reflect users' preferences.
First, we apply attention pooling over the complete GUI sequence to obtain the global visual-state representation:
\begin{equation}
\bar{x}
=
\operatorname{AttnPool}(c_{\text{img}}).
\label{eq:slow_pool}
\end{equation}

The pooled visual representation is then concatenated with $c_{\text{text}}$ and fed into a Transformer Encoder to obtain the long-term cross-modal representation $H_{\text{slow}}$ over $U$, $D$, $W$, and the complete GUI screenshot sequence:
\begin{equation}
H_{\text{slow}}
=
\mathcal{T}_{\text{slow}}
\left(
[c_{\text{text}};\bar{x}]
\right),
\label{eq:h_slow}
\end{equation}
where $\mathcal{T}_{\text{slow}}$ denotes the Transformer Encoder of the slow channel~\citep{Vaswani2017AttentionIA}.

The fast channel captures short-term user dynamics. Based on empirical observations, we select the most recent four representations from the projected GUI sequence $c_{\text{img}}$ and feed them into a Transformer Encoder to obtain the short-term dynamic representation $H_{\text{fast}}$:
\begin{equation}
H_{\text{fast}}
=
\mathcal{T}_{\text{fast}}
\left(
c_{\text{img}}[T-k+1:T]
\right),
\label{eq:h_fast}
\end{equation}
where $k=4$ and $\mathcal{T}_{\text{fast}}$ denotes the Transformer Encoder of the fast channel.

To enhance the information interaction between the fast and slow channels, MPP introduces a cross-attention mechanism, enabling the model to simultaneously consider users' long-term behavioral background and the instantaneous state of the current interface when determining the recommendation timing. Specifically, for any channel pair $(c,\bar{c}) \in \{(\mathrm{fast}, \mathrm{slow}), (\mathrm{slow}, \mathrm{fast})\}$, the representation of the current channel is used as the Query, while the representation of the other channel is used as the Key and Value, thereby obtaining the cross-channel context-enhanced representation:
\begin{equation}
\widetilde{H}_{c}
=
\mathrm{MHA}
\left(
Q = H_{c},
K = H_{\bar{c}},
V = H_{\bar{c}}
\right).
\label{eq:mpp_cross_attention}
\end{equation}

Attention pooling is then applied to the two enhanced representations, which are concatenated to obtain the fused representation $z \in \mathbb{R}^{2d}$.

The fused representation $z$ is fed into the trigger-gating head to obtain the probability of triggering proactive recommendation:
\begin{equation}
p_{trig} = \sigma(\mathrm{MLP}_{trig}(z)),
\label{eq:trigger_gate}
\end{equation}

\noindent where $\sigma$ denotes the sigmoid function. If $p_{trig}<\tau$, the sample is directly filtered out. Otherwise, it is passed to PAR for subsequent reasoning. The details are provided in Appendix~\ref{app:mpp_hyperparameter_sensitivity}.

To further compress the function pool and improve reasoning efficiency, we feed $z$ into the intent-scenario prediction head to obtain the probabilities of the 14 intent scenarios:

\begin{equation}
p_{scenario} = \sigma(\mathrm{MLP}_{scenario}(z)).
\label{eq:intent_prediction}
\end{equation}

We select the top-$K$ intent scenarios with the highest $p_{\mathrm{scenario}}$ and take the union of the functions associated with these scenarios to construct the compressed function pool $\tilde{F}$. Each function is associated with exactly one intent scenario, whereas one intent scenario corresponds to multiple functions. This operation reduces the function pool by approximately $68.25\%$. We set $K=5$ according to the analysis in Appendix~\ref{app:mpp_topk_sensitivity}.

\subsection{Proactive Agent Reasoner}
\label{sec:par}
For samples that MPP predicts as requiring recommendations, we input $U$, $D$, $W$, the complete GUI screenshot sequence, and the compressed candidate function set $\tilde{F}$ into PAR for structured reasoning and content generation. Since MPP serves as a lightweight pre-reasoning gate, PAR performs a final consistency verification before generating executable recommendations, mitigating potential gating errors from the upstream perceiver. To better predict users' intent, we design a multi-step reasoning process.

First, a UI summary is generated based on the GUI screenshots to understand the semantics and sequential information of the GUI screenshots. Then, combined with $U$, $D$, $W$, UI summary, and $\tilde{F}$, a thinking chain containing function selection is generated. Finally, PAR outputs the structured proactive recommendation results, including validated intervention decisions, selected function names, and corresponding parameters.

\begin{tcolorbox}[title=Reasoning Prompt $\mathcal{P}_r$, colback=gray!5!white, colframe=gray!75!black]
Given the user profile, device status, environmental information, screenshots, predicted scenes, and available functions, analyze the user’s current behavior and determine whether a recommendation is needed.
\\
1. ui\_summary: Summarize the current interface, key content, and the user’s operation stage.
\\
2. thinking: Analyze the user’s behavior, intent, task status, and whether a recommendation is needed; if needed, determine the recommendation function and parameters.
\\
3. recommendations: Output the recommendation instruction and function call; if no recommendation is needed, output “No recommendation”.
\end{tcolorbox}

\subsection{PRPF Training Objective}

Since the two stages play different roles, we train MPP and PAR separately, and use the training set of ProactiveMobile to train the two modules~\citep{proactivemobile}.

MPP is jointly optimized with two losses. Due to class imbalance, non-recommendation samples are oversampled to the same proportion as recommendation samples, and Focal Loss is used~\citep{8237586}. Given the ground-truth label $y_{\mathrm{trig}}$ and the predicted trigger probability $p_{\mathrm{trig}}$, where $y_{\mathrm{trig}}=1$ indicates triggering recommendation, the trigger-gate loss is denoted as $\mathcal{L}_{\mathrm{trig}}$:
\begin{equation}
p_t =
y_{\mathrm{trig}} p_{\mathrm{trig}}
+
(1-y_{\mathrm{trig}})(1-p_{\mathrm{trig}}),
\end{equation}
\begin{equation}
\mathcal{L}_{\mathrm{trig}}
=
-\alpha_t(1-p_t)^{\gamma}\log(p_t),
\end{equation}

\noindent where $p_t$ denotes the true-class predicted probability, $\gamma$ reduces the weight of well-classified samples, and $\alpha_t$ balances different categories.

For triggered samples, BCE loss is used for optimization, and we compute the function-scenario prediction loss $\mathcal{L}_{scenario}$:
\begin{equation}
\mathcal{L}_{\mathrm{scenario}}
=
\mathrm{BCE}
\left(
y_{\mathrm{scenario}},
p_{\mathrm{scenario}}
\right),
\end{equation}
where $y_{\mathrm{scenario}}$ denotes the corresponding ground-truth multi-label function-scenario vector, represented in multi-hot form. Finally, the joint training objective of MPP is defined as $\mathcal{L}_{\mathrm{MPP}}$:
\begin{equation}
\mathcal{L}_{\mathrm{MPP}}
=
\mathcal{L}_{\mathrm{trig}}
+
\lambda \mathcal{L}_{\mathrm{scenario}},
\label{eq:mpp_loss}
\end{equation}

\noindent where $\lambda$ is the weight of the function-scenario prediction loss.

To better train PAR, we further fine-tune Qwen3.5-9B using SFT and GRPO. We construct a total of 8,876 multimodal and text data samples for SFT on Qwen3.5-9B, and we present the detailed construction process in Appendix~\ref{app:sft_data_construction}. Thus, given the input $x_i$, the model generates the structured output $y_i$. The two types of data are uniformly trained using the standard autoregressive cross-entropy loss, and the SFT loss is defined as:
\begin{equation}
\resizebox{0.85\columnwidth}{!}{$
\mathcal{L}_{\mathrm{SFT}}
=
-\frac{1}{Z}
\sum_{\substack{i,j}}
m_{i,j}
\log P_{\theta}
\left(
y_{i,j}
\mid
x_i, y_{i,<j}
\right),
$}
\label{eq:sft_loss}
\end{equation}

\noindent where $y_{i,j}$ denotes the $j$-th token in the output of the $i$-th sample, and $y_{i,<j}$ denotes all tokens generated before the $j$-th token. $Z$ denotes the total number of valid supervised tokens.

While SFT teaches PAR the structured response format, it may still be unstable on boundary cases and imperfect in function selection or argument grounding. We therefore apply GRPO after SFT to stabilize no-intervention decisions and improve executable assistance generation~\citep{guo2025deepseek}.

We construct GRPO prompts from rollouts of the SFT-trained PAR, denoted as the reference policy $\pi_{\mathrm{ref}}$, and retain only rollout groups with non-trivial reward variance and intermediate correctness rates, leaving boundary no-intervention cases and intervention cases with disagreement in function selection or argument grounding. For each retained prompt $x_i$, GRPO samples $N$ rollouts $\{\hat{y}_{i,j}\}_{j=1}^{N}$ and forms the group-relative advantage
\begin{equation}
\hat{A}_{i,j}
=
\frac{R(\hat{y}_{i,j},x_i)-\mu_i}{\sigma_i+\delta},
\end{equation}
where $\mu_i$ and $\sigma_i$ are the mean and standard deviation of the rollout rewards, and $\delta$ is a small numerical constant. We optimize the policy with the clipped GRPO objective and a KL penalty to $\pi_{\mathrm{ref}}$. 

\noindent For a rollout $\hat{y}$, intervention outcomes are determined by whether the predicted and gold function sequences are empty. We use a structured reward
\begin{equation}
\resizebox{0.86\columnwidth}{!}{$
R(\hat{y},x)
=
R_{\mathrm{acc}}(\hat{y},x)
+
R_{\mathrm{fmt}}(\hat{y})
+
R_{\mathrm{cal}}(\hat{y},x),
$}
\end{equation}
where $R_{\mathrm{acc}}$ scores the intervention decision, function-name sequence matching, and argument grounding, while $R_{\mathrm{fmt}}$ scores compliance with the required output schema. The calibration term $R_{\mathrm{cal}}$ adjusts the reward across intervention outcomes: it anchors correct silence on no-intervention cases, keeps false interventions penalized even when they are well formed, rewards complete gold-function coverage, and discourages redundant function calls. Full reward-component definitions and calibration details are provided in Appendix~\ref{app:grpo_reward}.

\section{Experiments}
\subsection{Experimental Settings}
\label{sec:experimental_settings}

\paragraph{Training Details.}
The official ProactiveMobile training split contains 8,876 samples (see Appendix~\ref{app:dataset_supervision}). For MPP, we partition this split into internal-training and development subsets with an 8:2 ratio, train MPP on the internal-training subset, and select the inference hyperparameters $\tau$ and $K$ on the development subset with the trained checkpoint fixed. PAR is trained separately on the full official training split. 
The text and image encoders used in PRPF are BGE-small-zh-v1.5 and CLIP ViT-B-32\citep{xiao2024c,radford2021learning}.
The base model used by PRPF is Qwen3.5-9B. Specific experimental details are provided in Appendix~\ref{sec:implement_details}.

\paragraph{Baselines.}
To verify the effectiveness of PRPF, we select three representative types of models for comparison experiments. 
\textbf{Closed-source models:}
GPT-5.5, o3, Gemini-3.1-Pro, Claude-Opus-4.7, GLM-4.6V, Kimi-K2.5, and MiMo-V2.5~\citep{openai2026gpt55, openai2025o3, deepmind2026gemini31, anthropic2026claude47, hong2025glm, kimiteam2026kimik25visualagentic, xiaomi2026mimo25}. 
\textbf{Open-source models:}
TongUI-7B and Qwen3.5-9B~\citep{zhang2025tongui, qwen35blog}. \textbf{Proactive intelligence models:}
ProactiveMobile (7B) is the task-specific model introduced with the
ProactiveMobile benchmark. We additionally construct
UI-TARS-7B-DPO+Proactive and Qwen3.5-9B+Proactive by fine-tuning the
corresponding backbones on the ProactiveMobile training
split. The two reproduced baselines use the same SFT data, prompts,
output format, and training hyperparameters, differing only in their
backbones. All three baselines jointly perform the \emph{when} and
\emph{how} decisions within a unified end-to-end model, providing a
direct comparison with PRPF's explicitly decoupled architecture.

\begin{table*}[t]
\centering

\resizebox{\textwidth}{!}{
\begin{tabular}{lccccccccc}
\toprule
\multirow{2}{*}{\textbf{Model}} 
& \multicolumn{3}{c}{\textbf{Multimodal}} 
& \multicolumn{3}{c}{\textbf{Text}} 
& \multicolumn{3}{c}{\textbf{ALL}} \\
\cmidrule(lr){2-4} \cmidrule(lr){5-7} \cmidrule(lr){8-10}
& \textbf{Type-Acc}$\uparrow$ & \textbf{SR}$\uparrow$ & \textbf{FTR}$\downarrow$
& \textbf{Type-Acc}$\uparrow$ & \textbf{SR}$\uparrow$ & \textbf{FTR}$\downarrow$
& \textbf{Type-Acc}$\uparrow$ & \textbf{SR}$\uparrow$ & \textbf{FTR}$\downarrow$ \\
\midrule
GPT-5.5 
& \underline{40.34} & \textbf{18.40} & 33.48
& 55.09 & 49.02 & 19.24
& 47.71 & 33.69 & 23.84 \\

o3
& 39.63 & \underline{17.25} & 41.45
& 51.31 & 44.53 & 26.89
& 45.46 & 30.87 & 31.61 \\

Gemini-3.1-Pro
& 18.94 & 9.33 & 76.47
& 33.97 & 26.75 & 56.60
& 26.45 & 18.03 & 61.11 \\

Claude-Opus-4.7
& 37.23 & 15.83 & 41.03
& 55.91 & 48.69 & 18.81
& 46.56 & 32.24 & 26.11 \\

GLM-4.6V
& 36.25 & 11.74 & 74.45
& 29.98 & 20.19 & 67.04
& 33.12 & 15.96 & 69.33 \\

Kimi-K2.5
& 25.33 & 9.33 & 30.02
& 39.50 & 35.56 & 32.51
& 32.40 & 22.43 & 31.64 \\

MiMo-V2.5
& 32.10 & 11.63 & 39.45
& 39.08 & 34.54 & 37.06
& 35.58 & 23.07 & 37.87 \\

TongUI-7B
& 1.75 & 0.33 & \textbf{15.79}
& 8.15 & 8.04 & 18.58
& 4.95 & 4.18 & 18.10 \\

UI-TARS-7B-DPO+Proactive
& 30.90 & 8.52 & 63.13
& 14.17 & 9.90 & 64.67
& 22.54 & 9.21 & 63.84 \\

Qwen3.5-9B
& 8.30 & 2.95 & 71.52
& 8.15 & 5.69 & 77.13
& 8.22 & 4.32 & 75.30 \\

ProactiveMobile (7B)
& 33.68 & 15.61 & 23.91
& 56.84 & 26.04 & \underline{8.51}
& 45.25 & 20.82 & 13.76 \\

Qwen3.5-9B+Proactive
& 39.52 & 14.85 & 18.54
& \underline{59.52} & \underline{54.32} & 10.63
& \underline{49.51} & \underline{34.56} & \underline{13.49} \\

\textbf{PRPF (Ours)}
& \textbf{40.83} & 17.19 & \underline{17.99}
& \textbf{69.20} & \textbf{65.15} & \textbf{1.75}
& \textbf{55.00} & \textbf{41.15} & \textbf{7.21} \\
\bottomrule
\end{tabular}
}
\caption{Overall performance comparison on the ProactiveMobile test set. We report \(\mathrm{Type\text{-}Acc}\uparrow\), \(\mathrm{SR}\uparrow\), and \(\mathrm{FTR}\downarrow\) under multimodal, text, and overall settings. The best and second-best results are highlighted in bold and underlined, respectively. All scores are reported in percentage (\%).}
\label{tab:overall_performance}
\end{table*}
\paragraph{Metrics.}
Following ProactiveMobile, we adopt three core metrics: function-name sequence accuracy (Type-Acc), success rate (SR), and false trigger rate (FTR)~\citep{proactivemobile}. Type-Acc measures whether the predicted function-name sequence exactly matches the ground truth. SR evaluates the overall correctness of the prediction, while FTR quantifies the proportion of false proactive triggers among samples that do not require recommendation. The detailed calculation formulas are provided in Section~\ref{sec:expanded_results}.

\subsection{Overall Performance Analysis}

Table~\ref{tab:overall_performance} presents the overall performance comparison between PRPF and various baseline models on the ProactiveMobile test set. From these results, we make the following observations:

\paragraph{1. PRPF achieves the best overall performance.} Compared with the strongest fine-tuned baseline Qwen3.5-9B+Proactive, PRPF improves Type-Acc by 5.49 percentage points, improves SR by 6.59 percentage points, and reduces FTR by 6.28 percentage points. This suggests that, compared with standalone end-to-end VLMs, the two-stage PRPF better separates when to recommend from how to recommend, thereby improving recommendation success while suppressing false triggers.

\paragraph{2. FTR is significantly reduced.} From the table, we observe that models without task-specific fine-tuning often exhibit high FTR. For example, GPT-5.5 reaches 23.84\%. Fine-tuned models can better reduce false triggers, such as ProactiveMobile (7B), which reduces FTR to 13.76\%. In contrast, our method directly reduces FTR to 7.21\%, and even reduces it to 1.75\% in the text setting. The extremely low false trigger rate verifies the effectiveness of the two-stage training and inference of PRPF.

\subsection{Evaluation on ContextAgentBench}
\label{sec:cross_benchmark}

To further evaluate the generalization capability of PRPF beyond our primary benchmark, we additionally evaluate PRPF on ContextAgentBench, which assesses proactive timing and tool-use quality under a different task formulation, output schema, and data distribution.~\citep{yang2026contextagent}. The benchmark provides official training and test splits. For PRPF, we partition the official training split into training and development portions with a 9:1 ratio for training-time validation, while reserving the official test split for final evaluation. Both MPP and PAR are retrained on the training portion. PRPF uses a binary MPP gate for intervention decisions, followed by PAR, which scores candidate actions and generates tool-use outputs for triggered samples. GPT-5.5 and Qwen3.5-9B are evaluated under the ICL-ALL protocol used in the original benchmark.

As shown in Table~\ref{tab:contextagentbench}, PRPF achieves the best performance on all four metrics, including Acc-P, FD, RMSE, and Tool-F1. Compared with Qwen3.5-9B, PRPF improves Acc-P from 0.838 to 0.913, reduces FD from 0.028 to 0.008 and RMSE from 1.817 to 1.121, and improves Tool-F1 from 0.693 to 0.788. These results show that the decoupled perception-and-reasoning design remains effective under a different proactive-agent task and evaluation schema.

\begin{table}[t]
\centering
\small
\setlength{\tabcolsep}{4pt}
\renewcommand{\arraystretch}{1.15}
\begin{tabular}{lcccc}
\toprule
\textbf{Model}
& \textbf{Acc-P}$\uparrow$
& \textbf{FD}$\downarrow$
& \textbf{RMSE}$\downarrow$
& \textbf{Tool-F1}$\uparrow$ \\
\midrule
GPT-5.5
& \underline{0.908}
& \underline{0.020}
& \underline{1.556}
& \underline{0.737} \\

Qwen3.5-9B
& 0.838
& 0.028
& 1.817
& 0.693 \\

\textbf{PRPF (Ours)}
& \textbf{0.913}
& \textbf{0.008}
& \textbf{1.121}
& \textbf{0.788} \\
\bottomrule
\end{tabular}
\caption{Results on the ContextAgentBench test split. GPT-5.5 and Qwen3.5-9B are evaluated under the ICL-ALL protocol, while PRPF uses the trained MPP$\rightarrow$PAR pipeline. Higher values are better for Acc-P and Tool-F1, while lower values are better for FD and RMSE.}
\label{tab:contextagentbench}
\end{table}

\subsection{Ablation Study on Core Components}
\label{sec:ablation_study}

To further verify the effectiveness of each core component in PRPF, we conduct ablation studies on the same test set under the same settings, covering the following four module groups: \textbf{Overall Structure:} w/o MPP denotes removing MPP, where all samples directly enter PAR; w/o PAR denotes using Qwen3.5-9B for direct inference in the \emph{how} stage. \textbf{MPP Structure:} w/o Slow Channel denotes removing the slow channel from MPP, and w/o Fast Channel denotes removing the fast channel from MPP. \textbf{MPP Functionality:} w/o Compression denotes that MPP only provides the recommendation judgment, while the full function set is input into PAR; w/o Recommend denotes that MPP does not perform recommendation judgment and only provides the function compression results. \textbf{PAR Training Strategy:} w/o SFT denotes removing SFT from PAR, and w/o GRPO denotes removing GRPO from PAR.

\begin{table}[t]
\centering
\renewcommand{\arraystretch}{1.08}
\setlength{\tabcolsep}{3.2pt}
\resizebox{\columnwidth}{!}{
\begin{tabular}{@{}lcccccc@{}}
\toprule
\multirow{2}{*}{\textbf{Model}} 
& \multicolumn{2}{c}{\textbf{Multimodal}} 
& \multicolumn{2}{c}{\textbf{Text}} 
& \multicolumn{2}{c}{\textbf{ALL}} \\
\cmidrule(lr){2-3} \cmidrule(lr){4-5} \cmidrule(l){6-7}
& \textbf{SR~$\uparrow$} 
& \textbf{FTR~$\downarrow$} 
& \textbf{SR~$\uparrow$} 
& \textbf{FTR~$\downarrow$}  
& \textbf{SR~$\uparrow$} 
& \textbf{FTR~$\downarrow$} \\
\midrule
\textbf{Full PRPF (Ours)}
& \textbf{17.19} & \underline{17.99}
& \textbf{65.15} & \textbf{1.75}
& \textbf{41.15} & \textbf{7.21} \\

\multicolumn{1}{@{\hspace{0.8em}}l}{- w/o MPP}
& 15.56 & 22.01
& 61.16 & 4.22
& 38.33 & 10.38 \\

\multicolumn{1}{@{\hspace{0.8em}}l}{- w/o PAR}
& 6.06 & 60.64
& 60.89 & 3.33
& 33.44 & 18.46 \\

\multicolumn{1}{@{\hspace{0.8em}}l}{- w/o Slow Channel}
& 16.59 & 20.88
& 62.91 & 3.44
& 39.73 & 9.38 \\

\multicolumn{1}{@{\hspace{0.8em}}l}{- w/o Fast Channel}
& 16.54 & \textbf{17.73}
& 64.33 & \underline{2.02}
& 40.41 & \underline{7.32} \\

\multicolumn{1}{@{\hspace{0.8em}}l}{- w/o Compression}
& \underline{16.65} & 19.11
& \underline{64.50} & \underline{2.02}
& \underline{40.55} & 7.78 \\

\multicolumn{1}{@{\hspace{0.8em}}l}{- w/o Recommend}
& 15.88 & 21.28
& 62.75 & 3.87
& 39.29 & 9.74 \\

\multicolumn{1}{@{\hspace{0.8em}}l}{- w/o GRPO}
& 16.05 & 48.63
& 63.35 & 2.81
& 39.67 & 15.56 \\

\multicolumn{1}{@{\hspace{0.8em}}l}{- w/o SFT}
& 4.91 & 61.60
& 60.18 & 4.37
& 32.51 & 18.61 \\
\bottomrule
\end{tabular}
}
\caption{Ablation study of PRPF.}
\label{tab:ablation_core}
\end{table}
As shown in Table~\ref{tab:ablation_core}, MPP and PAR provide complementary benefits, as removing either module decreases ALL SR and increases FTR. Within MPP, the slow channel contributes more than the fast channel, since removing it leads to a larger drop in ALL SR and a greater increase in FTR. For MPP functionality, trigger gating plays a more important role than candidate function compression, as w/o Recommend causes more severe performance degradation than w/o Compression. For PAR training, SFT establishes the basic generation capability, whereas GRPO further improves the decision policy. Removing SFT substantially degrades both ALL SR and FTR, while removing GRPO mainly leads to a notable increase in ALL FTR.

\subsection{Efficiency Analysis}
\label{sec:efficiency_analysis}

We evaluate inference efficiency on the ProactiveMobile test set using three metrics: per-sample inference compute (TFLOPs), peak GPU memory (GB), and end-to-end latency (ms). We use the ProactiveMobile (7B) baseline introduced in Section~\ref{sec:experimental_settings} and the PRPF variants defined in Section~\ref{sec:ablation_study}. The efficiency comparison isolates the contributions of intervention gating and candidate-function compression by measuring their effects on the number of PAR invocations and the size of the function pool provided to PAR. Detailed routing and aggregation rules are provided in Appendix~\ref{app:efficiency_protocol}.

Compared with the single-stage ProactiveMobile (7B) baseline, our PRPF (9B) reduces compute by $69.3\%$ and end-to-end latency by $60.1\%$, at the cost of a $12.0\%$ increase in peak memory because PRPF runs a stronger 9B PAR backbone on the gate-accepted fraction. Both MPP components contribute to this saving, but in different ways. The intervention gate removes PAR calls for no-intervention observations, as reflected by PRPF w/o Compression a $24.6\%$ compute reduction and $29.8\%$ latency reduction; peak memory rises by $26.7\%$ because accepted contexts still use the full function pool on the 9B engine. By contrast, scenario-based function-pool filtering reduces the full pool of 63 API functions to no more than 20 functions on average, as reflected by PRPF w/o Recommend, which lowers the cost of every PAR call, reducing compute by $57.1\%$ and latency by $50.5\%$ with the same $12.0\%$ memory overhead. Thus, in the efficiency benchmark, candidate function-pool reduction has the larger standalone effect, while PRPF (9B) combines it with intervention gating so that heavy VLM-based assistance reasoning is invoked less often and with a shorter function-pool prompt when invoked.

\begin{figure}[t]
    \centering
    \includegraphics[width=\columnwidth]{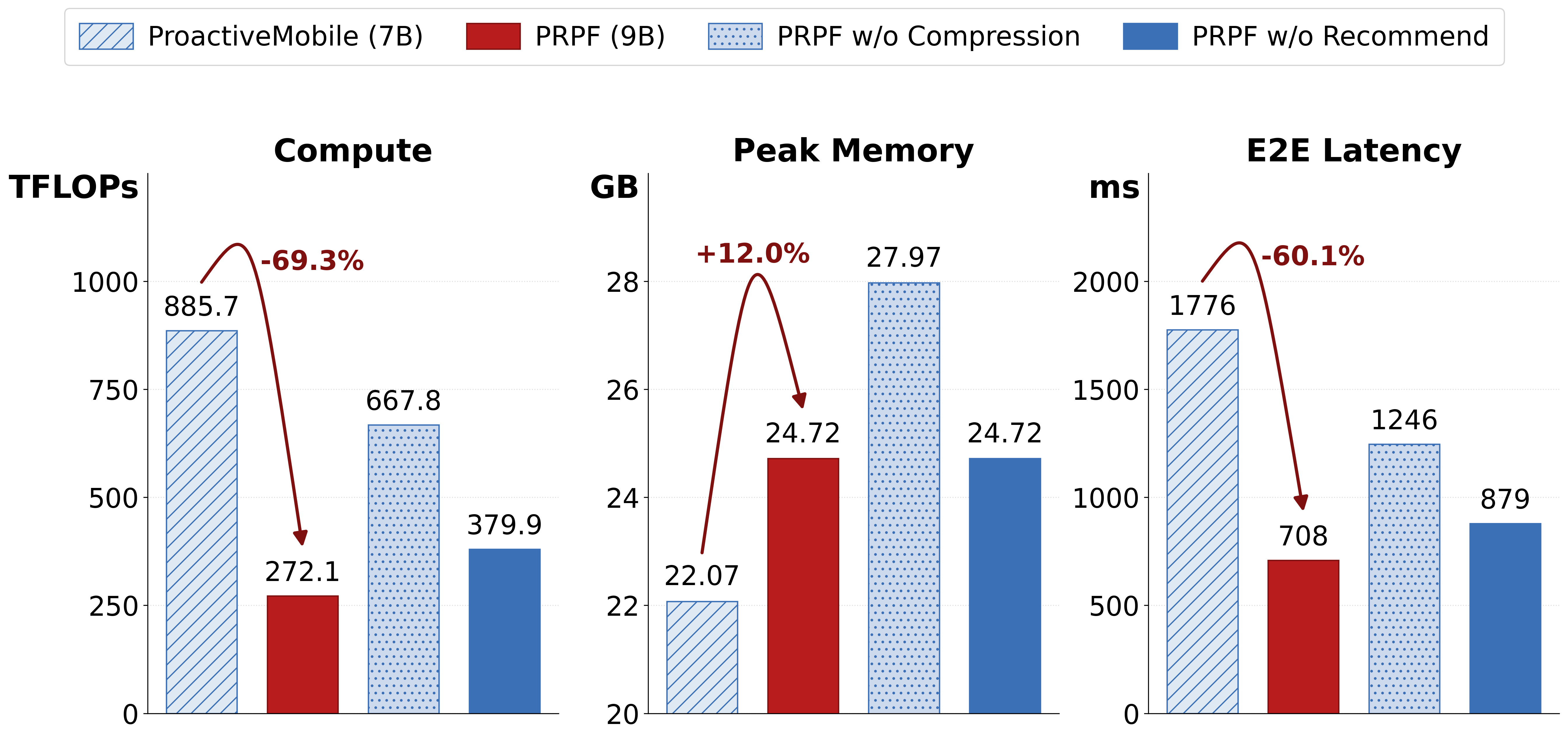}
    \caption{Inference efficiency comparison on ProactiveMobile. Lower values indicate better efficiency.}
    \label{fig:efficiency_analysis}
\end{figure}

\subsection{Analysis of MPP}
\label{sec:plug_and_play_mpp}

To evaluate the lightweight and plug-and-play properties of MPP, we consider three complementary comparisons. First, we compare ProactiveMobile (7B) with and without MPP, where MPP is inserted before the original proactive reasoner. Second, we perform the same comparison with GLM-4.6V to test whether MPP can be integrated with a different reasoner. In both comparisons, ``+ MPP'' denotes that MPP performs intervention gating and candidate-function compression before the downstream model generates the recommendation. Finally, we compare Qwen3.5-9B (SFT) + PAR with full PRPF. Here, Qwen3.5-9B (SFT) is a 9B model trained with the same SFT procedure as ProactiveMobile (7B) and used for intervention filtering and function selection, while PAR performs the subsequent recommendation generation. This comparison contrasts a larger perceptor with the lightweight MPP under the same two-stage setting. The results are reported in Table~\ref{tab:mpp_plug_and_play}.

\begin{table}[t]
\centering
\resizebox{\columnwidth}{!}{
\begin{tabular}{lcccccc}
\toprule
\multirow{2}{*}{\textbf{Model}} 
& \multicolumn{2}{c}{\textbf{Multimodal}} 
& \multicolumn{2}{c}{\textbf{Text}} 
& \multicolumn{2}{c}{\textbf{ALL}} \\
\cmidrule(lr){2-3} \cmidrule(lr){4-5} \cmidrule(lr){6-7}
& \textbf{SR~$\uparrow$} & \textbf{FTR~$\downarrow$} 
& \textbf{SR~$\uparrow$} & \textbf{FTR~$\downarrow$} 
& \textbf{SR~$\uparrow$} & \textbf{FTR~$\downarrow$} \\
\midrule
ProactiveMobile (7B) 
& 15.61 & 23.91 & 26.04 & 8.51 & 20.82 & 13.76 \\
ProactiveMobile (7B) + MPP 
& 12.61 & 21.95 & 62.20 & \textbf{1.60} & 37.38 & \underline{8.08} \\
GLM-4.6V 
& 11.74 & 74.45 & 20.19 & 67.04 & 15.96 & 69.33 \\
GLM-4.6V + MPP 
& 13.92 & 68.75 & 62.91 & 4.15 & 38.39 & 19.97 \\
Qwen3.5-9B (SFT) + PAR 
& \underline{17.08} & \underline{19.55} & \underline{62.58} & 3.55 & \underline{39.80} & 8.98 \\
\textbf{Full PRPF (Ours)} 
& \textbf{17.19} & \textbf{17.99} & \textbf{65.15} & \underline{1.75} & \textbf{41.15} & \textbf{7.21} \\
\bottomrule
\end{tabular}
}
\caption{Plug-and-play evaluation of MPP with different proactive reasoners.
The ``+ MPP'' variants prepend MPP for intervention gating and
candidate-function compression, while Qwen3.5-9B (SFT) as Perceptor + PAR uses
a fine-tuned 9B model as the perceptor and PAR as the downstream reasoner.}
\label{tab:mpp_plug_and_play}
\end{table}

The first two groups of comparisons demonstrate the plug-and-play property of MPP, showing that MPP becomes effective once integrated into different models. The third group verifies that MPP achieves better results than the fine-tuned Qwen3.5-9B while using far fewer parameters (0.1B vs. 9B).

\subsection{Case Study}
\label{sec:case_study}

\begin{figure}[t]
\centering
\includegraphics[width=\columnwidth]{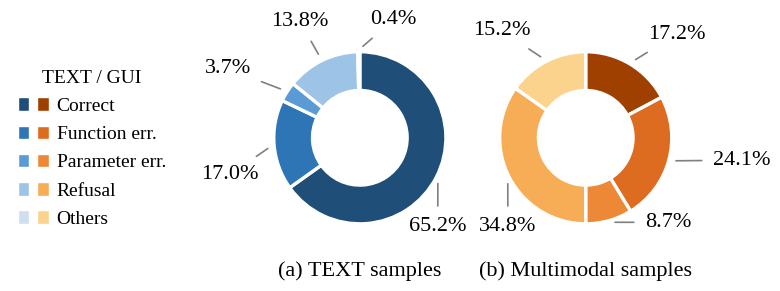}
\caption{Per-modality breakdown of PRPF outcomes on the ProactiveMobile test set under SR scoring.}
\label{fig:outcome_modality}
\end{figure}

To localize PRPF's failures, we partition every test sample into one of five mutually exclusive outcomes under SR scoring and report the per-modality breakdown in Figure~\ref{fig:outcome_modality}. PRPF achieves SR $=1$ on $65.2\%$ of TEXT samples but only $17.2\%$ of Multimodal samples; the multimodal gap is mainly caused by \emph{refusal} ($34.8\%$), where the system stays silent on samples that require recommendation, and \emph{function error} ($24.1\%$), where PAR selects an incorrect intent. A detailed qualitative case and a finer error decomposition (separating gate-side from PAR-side refusal, plus six sub-patterns of non-empty mismatch) are deferred to Appendix~\ref{app:failure}.

\section{Conclusion}

In this paper, we introduced PRPF, a Pre-Reasoning
Perception Framework for mobile proactive intelligence.
PRPF decouples when to recommend from how to recommend
by using MPP for lightweight pre-reasoning perception and
PAR for focused recommendation reasoning. On ProactiveMobile,
PRPF achieves 41.15\% overall success rate and 7.21\% false
trigger rate. It also reduces inference compute by 69.3\% and
end-to-end latency by 60.1\% over the ProactiveMobile (7B) baseline,
showing the lightweight efficiency of the proposed framework.
The consistent gains of MPP across different reasoners further
demonstrate its plug-and-play capability. 
The consistent improvements on ContextAgentBench further demonstrate the generalizability of PRPF across different proactive-agent task formulations, output schemas, and data distributions.
These results highlight
the value of separating lightweight intervention perception from
expensive recommendation reasoning in proactive mobile agents.

\section*{Limitations}

MPP is trained on the 14 high-level intent scenarios defined in the ProactiveMobile benchmark. While this design suffices for the current function pool, extending PRPF to new domains or a significantly larger API space would require re-training (or at least fine-tuning) the perceiver to learn new intent-scenario distributions and update the Top-$K$ filtering vocabulary. Future work could explore continual-learning or prompt-based adapters to mitigate this re-training cost.

Multimodal understanding remains a bottleneck. Despite the overall improvement, absolute success rates on multimodal tasks are still modest (17.19\% SR), and even the strongest baselines struggle to exceed 18\% SR on ProactiveMobile's GUI screenshot setting. This gap suggests that current vision-language models still lack the fine-grained interface understanding required for ProactiveMobile assistance. Closing this gap likely demands stronger visual grounding, higher-resolution GUI encoders, or larger-scale multimodal pre-training beyond the scope of this work.

\section*{Ethics Statement}
ProactiveMobile agents inherently require access to sensitive on-device signals, including GUI screenshots, user profiles, and interaction histories, which raises significant privacy concerns. We emphasize that all training and evaluation data used in this work are derived from the publicly available ProactiveMobile benchmark; no private user data or real-device traces were collected or used.
For real-world deployment, PRPF must enforce explicit user consent and strict adherence to the principle of least privilege to prevent unauthorized surveillance or data leakage. Beyond initial consent, users should retain granular control over which on-device signals the agent may access, with the ability to audit and modify these permissions at any time. An opt-out mechanism must allow users to disable proactive interventions globally or per-application without degrading passive query functionality. Furthermore, intervention personalization should enable users to customize the frequency, timing, and sensitivity of proactive triggers---for example, suppressing suggestions during specific activities or setting daily intervention limits---thereby preventing behavioral over-inference and ensuring the agent respects individual boundaries.

\bibliography{custom}

@inproceedings{proactiveagent,
  title={Proactive agent: Shifting llm agents from reactive responses to active assistance},
  author={Lu, Yaxi and Yang, Shenzhi and Qian, Cheng and Chen, Guirong and Luo, Qinyu and Wu, Yesai and Wang, Huadong and Cong, Xin and Zhang, Zhong and Lin, Yankai and others},
  booktitle={International Conference on Learning Representations},
  volume={2025},
  pages={47431--47457},
  year={2025}
}

@InProceedings{proactivemobile,
    author    = {Kong, Dezhi and Feng, Zhengzhao and Liang, Qiliang and Wang, Hao and Sun, Haofei and Yang, Changpeng and Li, Yang and Zhou, Peng and Nie, Shuai and Wang, Hongzhen and Zhou, Linfeng and Jia, Hao and Xu, Jiaming and Shi, Runyu and Huang, Ying},
    title     = {ProactiveMobile: A Comprehensive Benchmark for Boosting Proactive Intelligence On Mobile Devices},
    booktitle = {Proceedings of the IEEE/CVF Conference on Computer Vision and Pattern Recognition (CVPR)},
    month     = {June},
    year      = {2026},
    pages     = {27503-27513}
}

@misc{PARE,
      title={Proactive Agent Research Environment: Simulating Active Users to Evaluate Proactive Assistants}, 
      author={Deepak Nathani and Cheng Zhang and Chang Huan and Jiaming Shan and Yinfei Yang and Alkesh Patel and Zhe Gan and William Yang Wang and Michael Saxon and Xin Eric Wang},
      year={2026},
      eprint={2604.00842},
      archivePrefix={arXiv},
      primaryClass={cs.AI},
      url={https://arxiv.org/abs/2604.00842}, 
}

@article{vigil,
  author  = {Liu, Fengrui and He, Xiao and Zhang, Tieying},
  title   = {Help Without Being Asked: A Deployed Proactive Agent System for On-Call Support with Continuous Self-Improvement},
  journal = {arXiv preprint arXiv:2604.09579},
  year    = {2026}
}

@article{yang2026contextagent,
  title={Contextagent: Context-aware proactive llm agents with open-world sensory perceptions},
  author={Yang, Bufang and Xu, Lilin and Zeng, Liekang and Liu, Kaiwei and Jiang, Siyang and Lu, Wenrui and Chen, Hongkai and Jiang, Xiaofan and Xing, Guoliang and Yan, Zhenyu},
  journal={Advances in Neural Information Processing Systems},
  volume={38},
  pages={167509--167543},
  year={2026}
}

@inproceedings{zhang2025appagent,
  title={Appagent: Multimodal agents as smartphone users},
  author={Zhang, Chi and Yang, Zhao and Liu, Jiaxuan and Li, Yanda and Han, Yucheng and Chen, Xin and Huang, Zebiao and Fu, Bin and Yu, Gang},
  booktitle={Proceedings of the 2025 CHI Conference on Human Factors in Computing Systems},
  pages={1--20},
  year={2025}
}

@inproceedings{rawles2025androidworld,
  title={Androidworld: A dynamic benchmarking environment for autonomous agents},
  author={Rawles, Chris and Clinckemaillie, Sarah and Chang, Yifan and Waltz, Jonathan and Lau, Gabrielle and Fair, Marybeth and Li, Alice and Bishop, William and Li, Wei and Campbell-Ajala, Folawiyo and others},
  booktitle={International Conference on Learning Representations},
  volume={2025},
  pages={406--441},
  year={2025}
}

@misc{tang2026proagentbench,
      title={ProAgentBench: Evaluating LLM Agents for Proactive Assistance with Real-World Data}, 
      author={Yuanbo Tang and Huaze Tang and Tingyu Cao and Lam Nguyen and Anping Zhang and Xinwen Cao and Chunkang Liu and Wenbo Ding and Yang Li},
      year={2026},
      eprint={2602.04482},
      archivePrefix={arXiv},
      primaryClass={cs.HC},
      url={https://arxiv.org/abs/2602.04482}, 
}

@article{fu2026prism,
  title={PRISM: Festina Lente Proactivity--Risk-Sensitive, Uncertainty-Aware Deliberation for Proactive Agents},
  author={Fu, Yuxuan and Tan, Xiaoyu and Hao, Teqi and Zhan, Chen and Qiu, Xihe},
  journal={arXiv preprint arXiv:2602.01532},
  year={2026}
}

@article{xie2026pask,
  title={PASK: Toward Intent-Aware Proactive Agents with Long-Term Memory},
  author={Xie, Zhifei and Hu, Zongzheng and Ye, Fangda and Zhang, Xin and Chai, Haobo and Liu, Zihang and Wu, Pengcheng and Zhang, Guibin and Liao, Yue and Hu, Xiaobin and others},
  journal={arXiv preprint arXiv:2604.08000},
  year={2026}
}

@article{chai2026pira,
  title={PIRA-Bench: A Transition from Reactive GUI Agents to GUI-based Proactive Intent Recommendation Agents},
  author={Chai, Yuxiang and Tang, Shunye and Xiao, Han and Liu, Rui and Li, Hongsheng},
  journal={arXiv preprint arXiv:2603.08013},
  year={2026}
}

@article{guo2025deepseek,
  title={DeepSeek-R1 incentivizes reasoning in LLMs through reinforcement learning},
  author={Guo, Daya and Yang, Dejian and Zhang, Haowei and Song, Junxiao and Wang, Peiyi and Zhu, Qihao and Xu, Runxin and Zhang, Ruoyu and Ma, Shirong and Bi, Xiao and others},
  journal={Nature},
  volume={645},
  number={8081},
  pages={633--638},
  year={2025},
  publisher={Nature Publishing Group UK London}
}

@article{wang2024mobile,
  title={Mobile-agent: Autonomous multi-modal mobile device agent with visual perception},
  author={Wang, Junyang and Xu, Haiyang and Ye, Jiabo and Yan, Ming and Shen, Weizhou and Zhang, Ji and Huang, Fei and Sang, Jitao},
  journal={arXiv preprint arXiv:2401.16158},
  year={2024}
}

@inproceedings{deng2024mobile,
    title = "Mobile-Bench: An Evaluation Benchmark for {LLM}-based Mobile Agents",
    author = "Deng, Shihan  and
      Xu, Weikai  and
      Sun, Hongda  and
      Liu, Wei  and
      Tan, Tao  and
      Liu, Jianfeng  and
      Li, Ang  and
      Luan, Jian  and
      Wang, Bin  and
      Yan, Rui  and
      Shang, Shuo",
    editor = "Ku, Lun-Wei  and
      Martins, Andre  and
      Srikumar, Vivek",
    booktitle = "Proceedings of the 62nd Annual Meeting of the Association for Computational Linguistics (Volume 1: Long Papers)",
    month = aug,
    year = "2024",
    address = "Bangkok, Thailand",
    publisher = "Association for Computational Linguistics",
    url = "https://aclanthology.org/2024.acl-long.478/",
    doi = "10.18653/v1/2024.acl-long.478",
    pages = "8813--8831"
}

@inproceedings{cheng2024seeclick,
  title={Seeclick: Harnessing gui grounding for advanced visual gui agents},
  author={Cheng, Kanzhi and Sun, Qiushi and Chu, Yougang and Xu, Fangzhi and YanTao, Li and Zhang, Jianbing and Wu, Zhiyong},
  booktitle={Proceedings of the 62nd Annual Meeting of the Association for Computational Linguistics (Volume 1: Long Papers)},
  pages={9313--9332},
  year={2024}
}

@article{wu2025smoothing,
  title={Smoothing grounding and reasoning for mllm-powered gui agents with query-oriented pivot tasks},
  author={Wu, Zongru and Cheng, Pengzhou and Wu, Zheng and Ju, Tianjie and Zhang, Zhuosheng and Liu, Gongshen},
  journal={arXiv preprint arXiv:2503.00401},
  year={2025}
}

@article{mehrotra2025ishift,
  title={iSHIFT: Lightweight Slow-Fast GUI Agent with Adaptive Perception},
  author={Mehrotra, Sarthak and Rebbapragada, Sairam VC and Bonthu, Mani Hemanth Reddy and Balasubramanian, Vineeth N},
  journal={arXiv preprint arXiv:2512.22009},
  year={2025}
}

@article{tang2025think,
  title={Think twice, click once: Enhancing gui grounding via fast and slow systems},
  author={Tang, Fei and Shen, Yongliang and Zhang, Hang and Chen, Siqi and Hou, Guiyang and Zhang, Wenqi and Zhang, Wenqiao and Song, Kaitao and Lu, Weiming and Zhuang, Yueting},
  journal={arXiv preprint arXiv:2503.06470},
  year={2025}
}

@article{liu2026drs,
  title={DRS-GUI: Dynamic Region Search for Training-Free GUI Grounding},
  author={Liu, Yichao and Shen, Huawen and Yu, Liu and Liu, Shiyu and Chen, Zeyu and Zhou, Yu},
  journal={arXiv preprint arXiv:2605.15542},
  year={2026}
}

@article{ong2024routellm,
  title={Routellm: Learning to route llms with preference data},
  author={Ong, Isaac and Almahairi, Amjad and Wu, Vincent and Chiang, Wei-Lin and Wu, Tianhao and Gonzalez, Joseph E and Kadous, M Waleed and Stoica, Ion},
  journal={arXiv preprint arXiv:2406.18665},
  year={2024}
}

@inproceedings{xu2025learning,
  title={Learning to inference adaptively for multimodal large language models},
  author={Xu, Zhuoyan and Nguyen, Khoi Duc and Mukherjee, Preeti and Bagchi, Saurabh and Chaterji, Somali and Liang, Yingyu and Li, Yin},
  booktitle={Proceedings of the IEEE/CVF International Conference on Computer Vision},
  pages={3552--3563},
  year={2025}
}

@misc{qwen35blog,
    title = {Qwen3.5: Towards Native Multimodal Agents},
    url = {https://qwen.ai/blog?id=qwen3.5},
    author = {{Qwen Team}},
    month = {February},
    year = {2026}
}

@article{hurst2024gpt,
  title={{Gpt-4o} system card},
  author={Hurst, Aaron and Lerer, Adam and Goucher, Adam P and Perelman, Adam and Ramesh, Aditya and Clark, Aidan and Ostrow, AJ and Welihinda, Akila and Hayes, Alan and Radford, Alec and others},
  journal={arXiv preprint arXiv:2410.21276},
  year={2024}
}

@inproceedings{radford2021learning,
  title={Learning transferable visual models from natural language supervision},
  author={Radford, Alec and Kim, Jong Wook and Hallacy, Chris and Ramesh, Aditya and Goh, Gabriel and Agarwal, Sandhini and Sastry, Girish and Askell, Amanda and Mishkin, Pamela and Clark, Jack and others},
  booktitle={International conference on machine learning},
  pages={8748--8763},
  year={2021},
  organization={PmLR}
}

@misc{kimiteam2026kimik25visualagentic,
      title={Kimi K2.5: Visual Agentic Intelligence}, 
      author={{Kimi Team} and Tongtong Bai and Yifan Bai and Yiping Bao and S. H. Cai and Yuan Cao and Y. Charles and H. S. Che and others},
      year={2026},
      eprint={2602.02276},
      archivePrefix={arXiv},
      primaryClass={cs.CL},
      url={https://arxiv.org/abs/2602.02276}, 
}

@misc{qin2025uitarspioneeringautomatedgui,
      title={UI-TARS: Pioneering Automated GUI Interaction with Native Agents}, 
      author={Yujia Qin and Yining Ye and Junjie Fang and Haoming Wang and Shihao Liang and Shizuo Tian and Junda Zhang and others},
      year={2025},
      eprint={2501.12326},
      archivePrefix={arXiv},
      primaryClass={cs.AI},
      url={https://arxiv.org/abs/2501.12326}, 
}

@article{hong2025glm,
  title={{Glm-4.5 v} and {glm-4.1 v-thinking}: Towards versatile multimodal reasoning with scalable reinforcement learning},
  author={Hong, Wenyi and Yu, Wenmeng and Gu, Xiaotao and Wang, Guo and Gan, Guobing and Tang, Haomiao and Cheng, Jiale and Qi, Ji and Ji, Junhui and Pan, Lihang and others},
  journal={arXiv preprint arXiv:2507.01006},
  year={2025}
}

@article{zhang2025tongui,
  title={Tongui: Building generalized gui agents by learning from multimodal web tutorials},
  author={Zhang, Bofei and Shang, Zirui and Gao, Zhi and Zhang, Wang and Xie, Rui and Ma, Xiaojian and Yuan, Tao and Wu, Xinxiao and Zhu, Song-Chun and Li, Qing},
  journal={arXiv e-prints},
  pages={arXiv--2504},
  year={2025}
}

@techreport{openai2026gpt55,
  title={GPT-5.5 System Card},
  author={{OpenAI}},
  year={2026},
  url={https://openai.com/index/gpt-5-5-system-card/},
  institution={OpenAI}
}

@techreport{openai2025o3,
  title={OpenAI o3 and o4-mini System Card},
  author={{OpenAI}},
  year={2025},
  url={https://cdn.openai.com/pdf/2221c875-02dc-4789-800b-e7758f3722c1/o3-and-o4-mini-system-card.pdf},
  institution={OpenAI}
}

@techreport{deepmind2026gemini31,
  title={Gemini 3.1 Pro Model Card},
  author={{Google DeepMind}},
  year={2026},
  url={https://deepmind.google/models/model-cards/gemini-3-1-pro/},
  institution={Google DeepMind}
}

@inproceedings{xiao2024c,
  title={C-pack: Packed resources for general chinese embeddings},
  author={Xiao, Shitao and Liu, Zheng and Zhang, Peitian and Muennighoff, Niklas and Lian, Defu and Nie, Jian-Yun},
  booktitle={Proceedings of the 47th international ACM SIGIR conference on research and development in information retrieval},
  pages={641--649},
  year={2024}
}

@techreport{anthropic2026claude47,
  title={{Introducing Claude Opus 4.7}},
  author={{Anthropic}},
  year={2026},
  month={apr},
  url={https://www.anthropic.com/news/claude-opus-4-7},
  institution={Anthropic}
}

@misc{xiaomi2026mimo25,
  title        = {{MiMo-V2.5}},
  author       = {{Xiaomi MiMo}},
  year         = {2026},
  howpublished = {\url{https://mimo.mi.com/models/en-US/mimo-v2.5}},
  note         = {Accessed: 2026-08-28}
}

@article{comanici2025gemini,
  title={Gemini 2.5: Pushing the frontier with advanced reasoning, multimodality, long context, and next generation agentic capabilities},
  author={Comanici, Gheorghe and Bieber, Eric and Schaekermann, Mike and Pasupat, Ice and Sachdeva, Noveen and Dhillon, Inderjit and Blistein, Marcel and Ram, Ori and Zhang, Dan and Rosen, Evan and others},
  journal={arXiv preprint arXiv:2507.06261},
  year={2025}
}

@misc{anthropic2026claudesonnet46,
  title = {Claude Sonnet 4.6},
  author = {{Anthropic}},
  year = {2026},
  month = {feb},
  day = {17},
  howpublished = {\url{https://www.anthropic.com/news/claude-sonnet-4-6}},
  note = {Accessed: 2026-05-25}
}

@inproceedings{Gao2024CostEfficientLL,
  title={Cost-Efficient Large Language Model Serving for Multi-turn Conversations with CachedAttention},
  author={Bin Gao and Zhuomin He and Puru Sharma and Qingxuan Kang and Djordje Jevdjic and Junbo Deng and Xingkun Yang and Zhou Yu and Pengfei Zuo},
  booktitle={USENIX Annual Technical Conference},
  year={2024},
  url={https://api.semanticscholar.org/CorpusID:268793498}
}

@misc{yang2026proagentharnessingondemandsensory,
      title={ProAgent: Harnessing On-Demand Sensory Contexts for Proactive LLM Agent Systems in the Wild}, 
      author={Bufang Yang and Lilin Xu and Liekang Zeng and Yunqi Guo and Siyang Jiang and Wenrui Lu and Kaiwei Liu and Yixuan Li and Xiaofan Jiang and Guoliang Xing and Zhenyu Yan},
      year={2026},
      eprint={2512.06721},
      archivePrefix={arXiv},
      primaryClass={cs.AI},
      url={https://arxiv.org/abs/2512.06721}, 
}

@article{fawcett2006introduction,
  title={An introduction to ROC analysis},
  author={Fawcett, Tom},
  journal={Pattern recognition letters},
  volume={27},
  number={8},
  pages={861--874},
  year={2006},
  publisher={Elsevier}
}

@inproceedings{Vaswani2017AttentionIA,
  title={Attention is all you need},
  author={Ashish Vaswani and Noam Shazeer and Niki Parmar and Jakob Uszkoreit and Llion Jones and Aidan N. Gomez and Lukasz Kaiser and Illia Polosukhin},
  booktitle={Neural Information Processing Systems},
  year={2017},
  url={https://api.semanticscholar.org/CorpusID:13756489}
}

@INPROCEEDINGS{8237586,
  author={Lin, Tsung-Yi and Goyal, Priya and Girshick, Ross and He, Kaiming and Dollár, Piotr},
  booktitle={2017 IEEE International Conference on Computer Vision (ICCV)}, 
  title={Focal Loss for Dense Object Detection}, 
  year={2017},


  doi={10.1109/ICCV.2017.324}}

\appendix

\section{Experimental Settings}

\subsection{Implementation Details}
\label{sec:implement_details}
The training hyperparameters of MPP, PAR supervised fine-tuning (SFT) and GRPO optimization are summarized in Table~\ref{tab:training_hyperparameters}. All experiments are evaluated under the same setting, with the prompts, output format, and the list of available functions in the API pool kept consistent. PRPF is trained on NVIDIA H20 (96\,GB HBM3) GPUs, consuming approximately $2{,}817$ GPU-hours in total. 
\begin{table}[!t]
\centering
\small
\setlength{\tabcolsep}{6pt}
\renewcommand{\arraystretch}{1.15}
\begin{tabular}{ll}
\toprule
\textbf{Parameter} & \textbf{Value} \\
\midrule
\multicolumn{2}{l}{\textbf{MPP}} \\
\midrule
Optimizer & AdamW \\
Learning Rate & $1 \times 10^{-4}$ \\
Trigger-Gating Threshold & 0.78 \\
Candidate-Pool Size & $K=5$ \\
Batch Size & 8 \\
Maximum Epochs & 50 \\
Early-Stopping Patience & 10 \\
Hidden Dimension & 256 \\
Scenario Loss Weight & $\lambda=0.5$ \\
\midrule
\multicolumn{2}{l}{\textbf{PAR-SFT}} \\
\midrule
Optimizer & AdamW \\
Learning Rate & $1 \times 10^{-5}$ \\
LR Schedule & Cosine annealing \\
Warmup Ratio & 0.1 \\
Training Epochs & 5 \\
Batch Size & 16 \\
Gradient Accumulation Steps & 2 \\
\midrule
\multicolumn{2}{l}{\textbf{PAR-GRPO}} \\
\midrule
Optimizer & AdamW \\
Learning Rate & $5 \times 10^{-7}$ \\
KL Coefficient & 0.05 \\
Rollouts per Prompt & $N=8$ \\
Batch Size & 32 \\
Training Epochs & 1 \\
\bottomrule
\end{tabular}
\caption{Training hyperparameters for MPP, PAR supervised fine-tuning, and GRPO optimization.}
\label{tab:training_hyperparameters}
\vspace{-0.6em}
\end{table}



\subsection{Datasets}
\label{app:dataset_supervision}
ProactiveMobile is a comprehensive benchmark dataset for mobile proactive intelligence~\citep{proactivemobile}. It is designed to evaluate whether a model can proactively infer users' latent needs from contextual information on mobile devices before users explicitly issue requests, and generate executable function-call sequences.

The dataset is constructed around four types of mobile contexts, including user profiles, device states, world information, and behavioral trajectories. Specifically, user profiles describe users' long-term habits, preferences, and basic attributes; device states reflect the current device and environmental status; world information provides external background such as weather, time, and holidays; and behavioral trajectories characterize the continuous interaction process between users and mobile devices. Meanwhile, the dataset also constructs a fixed function pool containing 63 distinct composite API functions. In terms of data format, ProactiveMobile contains both text scenarios and multimodal scenarios, where text scenarios use textual descriptions of behavioral trajectories, while multimodal scenarios use consecutive GUI screenshot sequences. Detailed dataset statistics are reported in Table~\ref{tab:data_split}. In addition, considering the diversity of user needs in real-world mobile scenarios, ProactiveMobile adopts a multi-answer annotation mechanism, where the same context may have multiple reasonable intent--function sequence annotations. When multiple valid references are available, the prediction is matched against the best-matched valid gold sequence following the ProactiveMobile evaluation protocol.

\begin{table}[!t]
\centering
\resizebox{\columnwidth}{!}{
\begin{tabular}{cccccc}
\toprule
\textbf{Split} & \textbf{Data Type} & \textbf{Items} & \textbf{Intents} & \textbf{Images} 
& \textbf{Function Calls} \\
\midrule
\multirow{2}{*}{Train} 
& Multimodal & 4438 & 8977 & 32418 & 9964 \\
& Text       & 4438 & 4438 & -     & 8259 \\
\midrule
\multirow{2}{*}{Test}
& Multimodal & 1832 & 3711 & 14341 & 4173 \\
& Text       & 1828 & 2676 & -     & 2266 \\
\bottomrule
\end{tabular}
}
\caption{Statistics of the ProactiveMobile dataset, broken down by Train and Test splits and data modality. The table summarizes the composition of ProactiveMobile benchmark, including the number of items, intents, UI images, and total functions.}
\label{tab:data_split}
\end{table}

\subsection{SFT Data Construction}
\label{app:sft_data_construction}
We construct two types of SFT data according to the sample modality: multimodal SFT data and text SFT data.

\paragraph{Multimodal SFT data construction.}
We decompose the recommendation generation process into four types of supervision signals: interface-state understanding, function-selection constraints, reasoning-process construction, and final recommendation generation. For recommendation samples, we use Claude Sonnet 4.6 for generation~\citep{anthropic2026claudesonnet46}. Specifically, based on the GUI screenshot sequence, we first generate a UI Summary, which contains the current page stage, key interaction regions, and state changes. Then, combined with the user profile, device state, world information, and candidate functions, we generate the Thinking process containing Function Selection. The Function Selection includes candidate function-pool analysis, function-matching judgment, final function selection, and argument-source explanation. In this way, the model learns not only ``what to recommend'', but also ``based on what context to recommend'', ``why to recommend at this moment'', and ``how to select executable functions from the candidate function pool''~\citep{yang2026contextagent, proactivemobile}. For non-recommendation samples, we do not construct the Thinking process and directly output the result.

\paragraph{Text SFT data construction.}
We decompose the recommendation generation process into three types of supervision signals: function-selection constraints, reasoning-process construction, and final recommendation generation. Except for the generation of UI Summary, all generation steps are consistent with those of multimodal SFT data.

\subsection{GRPO Optimization Details}
\label{app:grpo_details}

We use GRPO as a post-SFT optimization stage for PAR~\citep{guo2025deepseek}. This appendix summarizes the rollout filtering, objective, and reward used for the GRPO stage.

\paragraph{Data construction.}
\label{app:grpo_data}
GRPO prompts are built from SFT-format prompts and rollouts of the SFT-trained PAR, which is used as the reference policy $\pi_{\mathrm{ref}}$. Let $\mathcal{D}_{\mathrm{roll}}$ denote the rollout set, where each case contains $N=8$ sampled outputs and rule-based reward scores. We retain only rollout groups with non-trivial disagreement:
\begin{equation}
\resizebox{0.85\columnwidth}{!}{$
\operatorname{std}(\{R_{i,j}\}_{j=1}^{N}) > 0.1,
\qquad
0.1 \leq \operatorname{CR}_i \leq 0.9,
$}
\end{equation}
Here $\operatorname{CR}_i$ is the rollout-level correctness rate under a case-specific intervention criterion. For intervention cases, correctness requires exceeding the TP-quality threshold used by the reward implementation, thereby filtering out mere triggering without sufficient function-name or argument correctness. For no-intervention cases, correctness means predicting an empty function-name sequence. Retained cases are weighted to emphasize function-correct but argument-weak cases and false-positive-prone no-intervention cases, then resized to $5{,}000$ examples with at least $30\%$ no-intervention records. During resizing, intervention records in the high-variance disagreement region, $\operatorname{std}_i>0.3$ and $0.1<\operatorname{CR}_i<0.7$, are preferentially preserved, followed by a $95{:}5$ stratified train/development split.


\paragraph{Reward.}
\label{app:grpo_reward}
Let $S(y)$ denote the function-name sequence extracted from output $y$. For the supervised target $y_i$, $S_i^\star=S(y_i)$ is the gold function-name sequence in the SFT response; for rollout $\hat{y}_{i,j}$, $\hat{S}_{i,j}=S(\hat{y}_{i,j})$ is the predicted sequence. When multiple gold recommendations are available, $S^\star$ denotes the best-matched gold sequence selected by exact sequence accuracy and then function-name F1. In the reward formulas below, we omit instance and rollout indices and write $\hat{S}$ and $S^\star$ for readability.

The four intervention outcomes are defined by whether $\hat{S}$ and $S^\star$ are empty:
\begin{equation}
\resizebox{0.85\columnwidth}{!}{$
\begin{aligned}
\mathbf{1}_{\mathrm{TP}} &= \mathbf{1}[|\hat{S}|>0 \land |S^\star|>0],&
\mathbf{1}_{\mathrm{TN}} &= \mathbf{1}[|\hat{S}|=0 \land |S^\star|=0],
\end{aligned}
$}
\end{equation}
\begin{equation}
\resizebox{0.85\columnwidth}{!}{$
\begin{aligned}
\mathbf{1}_{\mathrm{FP}} &= \mathbf{1}[|\hat{S}|>0 \land |S^\star|=0],&
\mathbf{1}_{\mathrm{FN}} &= \mathbf{1}[|\hat{S}|=0 \land |S^\star|>0].
\end{aligned}
$}
\end{equation}

The rollout reward is decomposed as
\begin{equation}
\resizebox{0.85\columnwidth}{!}{$
R(\hat{y},x)
= R_{\mathrm{acc}}(\hat{y},x)
+
R_{\mathrm{fmt}}(\hat{y})
+
R_{\mathrm{cal}}(\hat{y},x),
$}
\end{equation}
\begin{equation}
\resizebox{0.85\columnwidth}{!}{$
R_{\mathrm{acc}}
= w_T R_{\mathrm{trig}}
+
(w_F R_{\mathrm{func}}+w_A R_{\mathrm{args}})
\mathbf{1}_{\mathrm{TP}},
$}
\end{equation}
\begin{equation}
R_{\mathrm{fmt}}
= w_M R_{\mathrm{schema}},
\end{equation}
\begin{equation}
\resizebox{0.85\columnwidth}{!}{$
R_{\mathrm{cal}}
= b_{\mathrm{TN}}\mathbf{1}_{\mathrm{TN}}
+
(R_{\mathrm{fp}}+p_{\mathrm{FP}})\mathbf{1}_{\mathrm{FP}}
+
(b_{\mathrm{cov}}+p_{\mathrm{over}})\mathbf{1}_{\mathrm{TP}}.
$}
\end{equation}

$R_{\mathrm{trig}}$ assigns quadrant-level intervention credit to TP/TN/FP/FN outcomes. $R_{\mathrm{func}}$ combines function-name set-F1 and longest-common-subsequence order consistency. $R_{\mathrm{args}}$ combines must-fill argument completeness and typed value similarity. $R_{\mathrm{schema}}$ scores output schema compliance, including tag order, function-field presence, JSON validity, and function-call structure. The calibration terms reward correct silence, keep false interventions net-penalized, reward full gold-function coverage, and penalize redundant function calls. Unparseable outputs receive only schema credit.

\begin{figure}[t]
    \centering
    \includegraphics[width=\columnwidth]{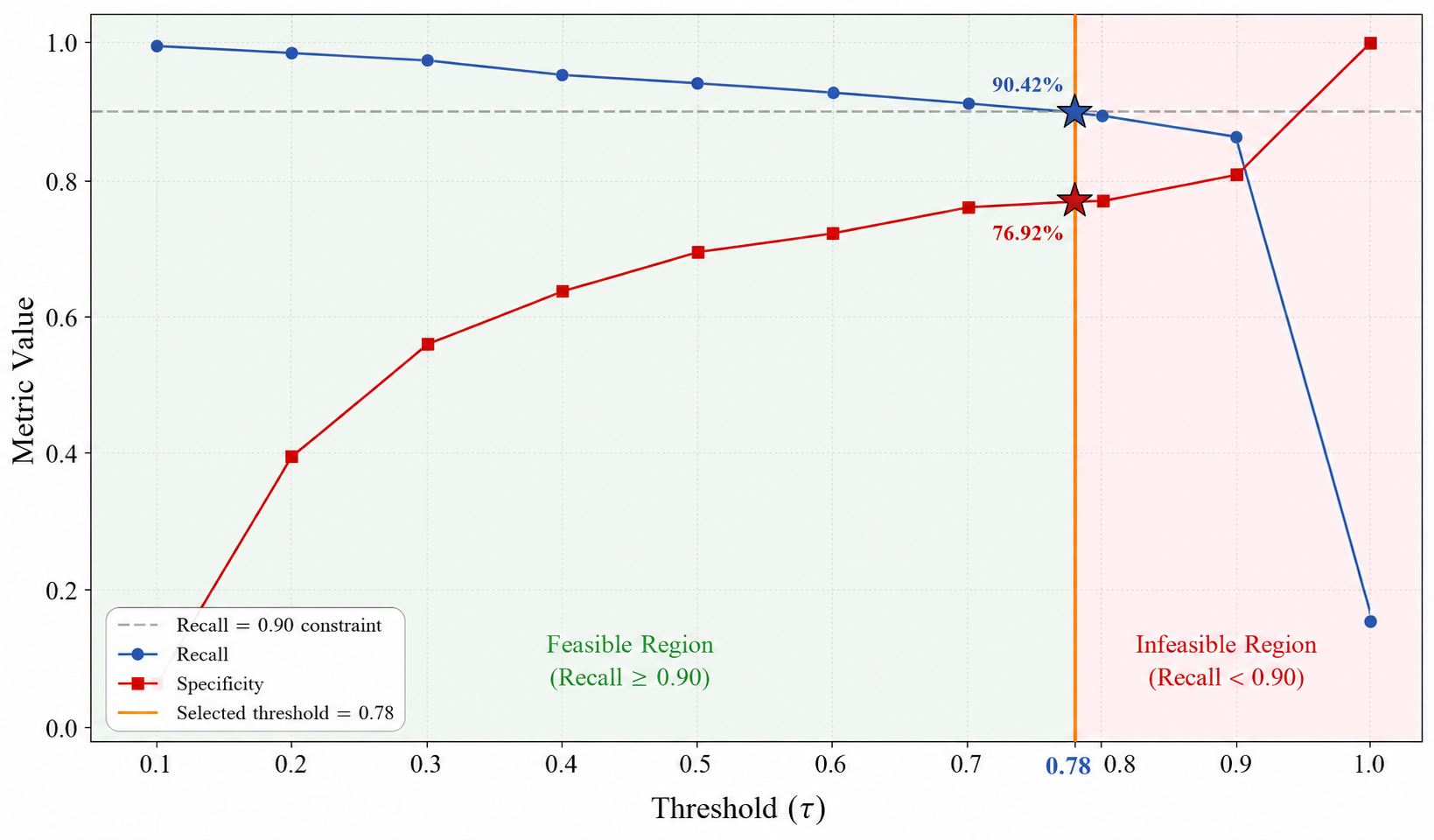}
    \caption{Sensitivity analysis of trigger-gate performance under different thresholds $\tau$.}
    \label{fig:threshold_curve}
\end{figure}

We set $w_T=2.0$, $w_F=3.0$, $w_A=2.5$, and $w_M=0.5$, with $b_{\mathrm{TN}}=6.0$, $p_{\mathrm{FP}}=-1.5$, $b_{\mathrm{cov}}=0.2$, and $\lambda_o=0.6$ for the over-prediction penalty. These constants are calibrated on development rollouts to preserve the intended ordering from malformed outputs, to valid no-intervention outputs, to partially correct TP outputs, and finally to fully correct executable recommendations. Our empirical ablations focus on the MPP/PAR architecture rather than treating reward constants as a separate contribution.




\section{Additional Experimental Results}

\subsection{MPP Hyperparameter Selection and Sensitivity Analysis}
\label{app:mpp_hyperparameter_sensitivity}
Following the protocol described in Section~\ref{sec:experimental_settings}, we use the fixed MPP checkpoint trained on the internal-training subset and select $\tau$ and $K$ on the corresponding development subset. The selected values are then used unchanged on the official test split.

\paragraph{Trigger-Gating Threshold $\tau$.}
\label{app:mpp_threshold_selection}

The main hyperparameter affecting the effectiveness of MPP is the threshold $\tau$ of the trigger-gating head, which directly determines how many samples need to be filtered. Since the goal of MPP is to filter out a large number of non-recommendation samples while retaining as many recommendation samples as possible, we mainly select two evaluation metrics: Recall and Specificity. Recall measures the proportion of recommendation samples that are correctly retained, i.e., avoiding missed recommendations, while Specificity measures the proportion of non-recommendation samples that are correctly filtered, i.e., effectively suppressing false triggers:
\begin{equation}
\resizebox{0.85\columnwidth}{!}{$
\mathrm{Recall}=\frac{\mathrm{TP}}{\mathrm{TP}+\mathrm{FN}}, 
\mathrm{Specificity}=\frac{\mathrm{TN}}{\mathrm{TN}+\mathrm{FP}}.
$}
\end{equation}

Following the standard idea of ROC operating-point selection~\citep{fawcett2006introduction}, we formulate threshold selection as a constrained optimization problem: under the constraint of $\mathrm{Recall} \geq 0.90$, i.e., the missed recommendation rate does not exceed $10\%$, we maximize Specificity. As shown in Figure~\ref{fig:threshold_curve}, we perform a grid search on the development subset over thresholds $\tau \in [0.05, 0.95]$ with a step size of $0.01$. As $\tau$ increases, Recall monotonically decreases while Specificity monotonically increases, forming a trade-off curve between the two metrics. We select $\tau=0.78$ as the deployment threshold, achieving $\mathrm{Recall}=90.42\%$ and $\mathrm{Specificity}=76.92\%$ on the dev split, and freeze it for all subsequent experiments. This means that the gate can filter out approximately $77\%$ of non-recommendation samples while retaining more than $90\%$ of samples that truly require recommendations. Among the thresholds satisfying the recall constraint, $\tau=0.78$ provides the strongest filtering capability on the dev split.

\paragraph{Top-$K$ Intent Scenarios.}
\label{app:mpp_topk_sensitivity}
With $\tau$ fixed at the value selected above, we sweep the number of selected intent scenarios, $K \in \{1,\ldots,10\}$ on the same development subset. We report development-set Top-$K$ Recall and end-to-end SR in Table~\ref{tab:topk_sensitivity}. Top-$K$ intent-scenario Recall counts an instance as recalled if the selected intent scenarios cover all intent scenarios associated with at least one ground-truth function sequence.
\begin{table*}[t]
\centering
\small
\renewcommand{\arraystretch}{1.20}
\begin{tabular*}{\textwidth}{@{\extracolsep{\fill}}lcccccccccc}
\toprule
$K$ & 1 & 2 & 3 & 4 & \textbf{5} & 6 & 7 & 8 & 9 & 10 \\
\midrule
Dev Top-$K$ Recall (\%) & 74.92 & 85.19 & 90.66 & 93.69 & \textbf{95.52} & 97.02 & 97.84 & 98.47 & 98.93 & 99.37 \\
Dev SR (\%)             & 39.54 & 39.89 & 40.90 & 40.19 & \textbf{41.15} & 40.87 & 40.49 & 40.98 & 40.55 & 40.11 \\
\bottomrule
\end{tabular*}
\caption{Sensitivity to the number of selected intent scenarios $K$ on the development set. The bolded column ($K=5$) marks the saturation point of Top-$K$ Recall, at which SR also peaks; we adopt it as the default.}
\label{tab:topk_sensitivity}
\end{table*} \textbf{$K=5$ provides the best development-set trade-off.}
On the held-out development subset, Top-$K$ Recall reaches 95.52\% at $K=5$ and improves by only an additional 3.85 percentage points when $K$ is increased from 5 to 10. Meanwhile, development-set SR reaches its highest value, 41.15\%, at $K=5$ and does not improve with larger sets of selected intent scenarios. This non-monotonic behavior is consistent with distractor interference: once the relevant functions are reliably covered, additional candidates increase the prompt and candidate-scoring burden without consistently improving the final prediction. We therefore select $K=5$ by jointly considering candidate coverage, end-to-end development SR, and function-pool compactness. The selected value is frozen and used unchanged in all final test-set evaluations.

\subsection{Detailed Results and Experimental Analysis of MPP}
\label{app:mpp_detailed_results}
For multimodal data, the processing procedure of MPP has been described in the main text. For text data, we only need to replace the GUI screenshot sequence with Trace Text, while keeping all other steps consistent.

\begin{table}[t]
\centering
\resizebox{\columnwidth}{!}{
\begin{tabular}{lccc}
\toprule
 & \textbf{Pred. Pos.} & \textbf{Pred. Neg.} & \textbf{Total} \\
\midrule
\textbf{Actual Pos.} 
& 2171 & 239 & 2410 \\
\textbf{Actual Neg.} 
& 169 & 1081 & 1250 \\
\midrule
\textbf{Total} 
& 2340 & 1320 & 3660 \\
\bottomrule
\end{tabular}
}
\caption{Trigger-decision confusion matrix of MPP on the ProactiveMobile test set. Pos. and Neg. denote recommendation and non-recommendation, respectively.}
\label{tab:mpp_confusion_matrix}
\end{table}
The main experiments use the seed-42 MPP checkpoint with the fixed threshold $\tau=0.78$. The confusion matrix is shown in Table~\ref{tab:mpp_confusion_matrix}. MPP filters out 1,320 samples, of which 1,081 are actual non-recommendation cases, corresponding to a negative predictive value of 81.89\% ($1{,}081/1{,}320$) for the filtered subset. Moreover, a portion of the 169 cases will also be correctly handled in the subsequent PAR stage.


For the five-seed stability analysis, each seed independently selects its own threshold on its corresponding development subset using the same recall-constrained rule. The threshold $0.59\pm0.24$ reported in Table~\ref{tab:mpp_seed_analysis} is the mean $\pm$ standard deviation of five seed-specific thresholds. Table~\ref{tab:mpp_seed_analysis} reports the five-seed mean $\pm$ standard deviation. The results indicate that the trigger decision is stable and reliable. The filtering capability is also effective: Specificity reaches 83.81--86.48\%, which means that MPP can filter out more than 83\% of non-recommendation samples, significantly reducing the computational burden of the downstream decision model. 
\begin{equation}
\mathrm{Accuracy} =
\frac{TP + TN}{TP + TN + FP + FN}
\end{equation}
\begin{equation}
\mathrm{Precision} =
\frac{TP}{TP + FP}
\end{equation}
\begin{equation}
\mathrm{F1} =
\frac{2 \times \mathrm{Precision} \times \mathrm{Recall}}
{\mathrm{Precision} + \mathrm{Recall}}
\end{equation}
\begin{equation}
\mathrm{Intent\ Top\text{-}5\ Acc} =
\frac{\sum_{c=1}^{C} N_c^{\mathrm{correct}}}
{\sum_{c=1}^{C} N_c},
\end{equation}

\noindent where $C$ denotes the total number of scenario categories, $N_c^{\mathrm{correct}}$ denotes the number of samples in the $c$-th scenario category that are correctly hit by the Top-5 prediction, and $N_c$ denotes the total number of samples in the $c$-th scenario category.

\begin{table*}[t]
\centering
\resizebox{\textwidth}{!}{
\begin{tabular}{ccccccc}
\toprule
\textbf{Threshold} & \textbf{Accuracy} & \textbf{Recall} & \textbf{Specificity} & \textbf{Precision} & \textbf{F1} & \textbf{Intent Top-5 Acc} \\
\midrule
$0.59 \pm 0.24$ & $87.99 \pm 0.78$ & $90.16 \pm 0.11$ & $83.81 \pm 2.45$ & $91.49 \pm 1.17$ & $90.82 \pm 0.53$ & $87.95 \pm 3.48$ \\
\bottomrule
\end{tabular}
}
\caption{Performance analysis of MPP under different random seeds.}
\label{tab:mpp_seed_analysis}
\end{table*}

\subsection{Expanded Results with Granular Metrics}
\label{sec:expanded_results}
\paragraph{Metric Calculation.}
Given a test set $\mathcal{D}$, let $S(y)$ denote the function-name sequence extracted from an output $y$. 
For a model prediction $\hat{y}$, we denote the predicted function-name sequence as
$
\hat{S}=S(\hat{y})
$.

When multiple valid ground-truth recommendations are available,
$
\mathcal{G}
=
\{S^{\star}_{1},\ldots,S^{\star}_{K}\}
$
denotes the set of all valid gold function-name sequences.
Each sequence consists of ordered function names.

For metrics that require matching against a single reference, $S^\star$ denotes the best-matched gold function-name sequence selected from $\mathcal{G}$ by maximizing exact sequence accuracy and then function-name F1.

For SR evaluation, the associated function arguments are additionally considered by the LLM judge.

Following ProactiveMobile, the calculation formulas for Type-Acc, SR, and FTR are shown in Eq.~\eqref{eq:type_acc}, Eq.~\eqref{eq:sr}, and Eq.~\eqref{eq:ftr}, respectively.

\begin{equation}
\resizebox{0.85\columnwidth}{!}{$
\mathrm{Type\text{-}Acc}
=
\frac{1}{|\mathcal{D}|}
\sum_{i \in \mathcal{D}}
\mathbb{I}\!\left[
\exists S^\star \in \mathcal{G},\,
\hat{S}=S^\star
\right],
$}
\label{eq:type_acc}
\end{equation}

\noindent where $|\mathcal{D}|$ denotes the number of test instances, $\mathcal{G}$ denotes the set of valid gold function-name sequences, and $\mathbb{I}[\cdot]$ is the indicator function that returns 1 if the condition is satisfied and 0 otherwise.

\begin{equation}
\resizebox{0.85\columnwidth}{!}{$
\mathrm{SR}
=
\frac{1}{|\mathcal{D}|}
\sum_{i \in \mathcal{D}}
\mathbb{I}\!\left[
\exists S^\star \in \mathcal{G},\,
\hat{y}
\equiv_{\mathrm{func}}
y^\star
\right],
$}
\label{eq:sr}
\end{equation}

\noindent where $y^\star$ denotes the ground-truth response corresponding to $S^\star$, and $\equiv_{\mathrm{func}}$ denotes functional equivalence judged by the evaluation protocol.

Following ProactiveMobile, we employ Gemini-2.5-Pro as the LLM judge to evaluate functional equivalence~\citep{comanici2025gemini}. Specifically, when the predicted output $\hat{y}$ and a ground-truth response $y^\star$ are functionally equivalent, i.e., the selected functions and their corresponding arguments are semantically consistent, the SR score of this instance is assigned as 1; otherwise, the SR score is 0.

\begin{equation}
\mathrm{FTR}
=
\frac{
\sum_{i \in \mathcal{D}}
\mathbb{I}
\left[
\mathcal{G}=\emptyset
\land
\hat{S} \neq \emptyset
\right]
}{
\sum_{i \in \mathcal{D}}
\mathbb{I}
\left[
\mathcal{G}=\emptyset
\right]
},
\label{eq:ftr}
\end{equation}

\noindent where $\mathcal{G}=\emptyset$ denotes that the current instance does not require proactive recommendation, while $\hat{S} \neq \emptyset$ indicates that the model incorrectly triggers a recommendation.

\begin{table*}[!htbp]
\centering
\tiny
\setlength{\tabcolsep}{1.2pt}
\renewcommand{\arraystretch}{1.1}
\resizebox{\textwidth}{!}{%
\begin{tabular}{@{}ll*{18}{c}@{}}
\toprule
\multirow{2}{*}{\textbf{Difficulty}} 
& \multirow{2}{*}{\textbf{Model}} 
& \multicolumn{6}{c}{\textbf{Multimodal}} 
& \multicolumn{6}{c}{\textbf{Text}} 
& \multicolumn{6}{c}{\textbf{ALL}} \\
\cmidrule(lr){3-8} \cmidrule(lr){9-14} \cmidrule(lr){15-20}
& & \textbf{Type-Acc$\uparrow$} & \textbf{SR$\uparrow$} & \textbf{FTR$\downarrow$} & \textbf{F1$\uparrow$} & \textbf{P$\uparrow$} & \textbf{R$\uparrow$} 
& \textbf{Type-Acc$\uparrow$} & \textbf{SR$\uparrow$} & \textbf{FTR$\downarrow$} & \textbf{F1$\uparrow$} & \textbf{P$\uparrow$} & \textbf{R$\uparrow$} 
& \textbf{Type-Acc$\uparrow$} & \textbf{SR$\uparrow$} & \textbf{FTR$\downarrow$} & \textbf{F1$\uparrow$} & \textbf{P$\uparrow$} & \textbf{R$\uparrow$} \\
\midrule

\multirow{13}{*}{\textbf{L1}} 
& GPT-5.5 & \underline{52.54} & 22.03 & 57.14 & 61.72 & \underline{63.14} & 62.15 & 83.78 & 81.08 & 7.76 & 87.65 & 88.22 & 87.90 & 74.01 & 62.60 & 13.36 & 79.53 & \underline{80.37} & \underline{79.84} \\
& o3 & \textbf{53.39} & \underline{22.88} & 74.07 & 60.03 & 61.44 & 59.75 & 80.70 & 76.45 & 14.16 & 82.05 & 82.63 & 81.79 & 72.15 & 59.68 & 20.73 & 75.16 & 76.00 & 74.89 \\
& Gemini-3.1-Pro & 36.44 & 14.41 & 93.33 & 47.85 & 48.02 & 49.58 & 58.69 & 55.60 & 38.21 & 63.35 & 63.45 & 64.03 & 51.72 & 42.71 & 41.85 & 58.50 & 58.62 & 59.51 \\
& Claude-Opus-4.7 & 48.31 & \textbf{23.73} & 76.00 & \underline{63.05} & \underline{63.14} & \underline{65.11} & \underline{86.87} & 82.24 & 8.15 & \underline{88.10} & \underline{88.61} & \underline{87.97} & \underline{74.80} & \underline{63.93} & 15.04 & \textbf{80.26} & \textbf{80.64} & \textbf{80.81} \\
& GLM-4.6V & 50.85 & \textbf{23.73} & 91.30 & \textbf{63.14} & \textbf{64.41} & 63.84 & 43.63 & 37.84 & 58.57 & 45.61 & 46.08 & 45.62 & 45.89 & 33.42 & 61.80 & 51.10 & 51.81 & 51.33 \\
& Kimi-K2.5 & 32.20 & 13.56 & 40.63 & 49.69 & 47.02 & 59.61 & 67.95 & 64.87 & 24.54 & 70.21 & 69.88 & 71.24 & 56.76 & 48.81 & 26.61 & 63.78 & 62.73 & 67.60 \\
& MiMo-V2.5 & 50.85 & 13.56 & 48.28 & 62.54 & 62.01 & \textbf{65.40} & 59.46 & 57.53 & 32.55 & 63.26 & 62.77 & 65.19 & 56.76 & 43.77 & 34.44 & 63.03 & 62.53 & 65.25 \\
& TongUI-7B & 1.70 & 0.00 & \textbf{0.00} & 1.70 & 1.70 & 1.70 & 8.11 & 8.11 & 32.26 & 8.11 & 8.11 & 8.11 & 6.10 & 5.57 & 30.30 & 6.10 & 6.10 & 6.10 \\
& UI-TARS-7B-DPO+Proactive & 36.44 & 6.78 & 75.00 & 43.08 & 44.92 & 42.23 & 21.24 & 18.92 & 53.85 & 22.27 & 22.78 & 22.01 & 26.00 & 15.12 & 57.66 & 28.78 & 29.71 & 28.34 \\
& Qwen3.5-9B & 14.41 & 5.09 & 54.55 & 17.94 & 18.08 & 18.64 & 5.41 & 4.63 & 86.49 & 6.98 & 7.14 & 6.89 & 8.22 & 4.78 & 82.35 & 10.41 & 10.57 & 10.57 \\
& ProactiveMobile (7B) & 32.20 & 10.17 & 42.31 & 38.84 & 40.25 & 38.56 & 82.24 & 37.07 & 6.82 & 83.45 & 83.46 & 83.59 & 66.58 & 28.65 & 10.57 & 69.49 & 69.94 & 69.50 \\
& Qwen3.5-9B+Proactive & 38.98 & 5.09 & \underline{21.62} & 42.09 & 43.22 & 41.53 & \underline{86.87} & \underline{84.17} & \underline{3.65} & 87.65 & 87.65 & 87.65 & 71.88 & 59.42 & \underline{6.25} & 73.39 & 73.74 & 73.21 \\
& \textbf{PRPF (Ours)} & 43.22 & 13.56 & 32.26 & 50.00 & 50.85 & 50.00 & \textbf{92.28} & \textbf{90.35} & \textbf{0.90} & \textbf{93.31} & \textbf{93.44} & \textbf{93.44} & \textbf{76.92} & \textbf{66.31} & \textbf{4.74} & \underline{79.75} & 80.11 & \underline{79.84} \\
\midrule

\multirow{13}{*}{\textbf{L2}} 
& GPT-5.5 & \underline{42.41} & \textbf{19.58} & 38.24 & \underline{55.91} & \textbf{55.85} & \underline{58.62} & 64.14 & 58.51 & 20.33 & 71.89 & 72.81 & 72.17 & 54.08 & 40.48 & 24.27 & \underline{64.49} & \underline{64.96} & 65.89 \\
& o3 & 41.60 & \underline{19.09} & 54.76 & 53.93 & 55.49 & 54.30 & 58.09 & 51.06 & 31.26 & 64.28 & 65.68 & 63.98 & 50.45 & 36.25 & 36.26 & 59.48 & 60.96 & 59.49 \\
& Gemini-3.1-Pro & 21.04 & 10.28 & 81.25 & 39.64 & 37.65 & 45.73 & 34.60 & 28.27 & 62.28 & 47.10 & 47.29 & 49.35 & 28.32 & 19.94 & 64.88 & 43.64 & 42.83 & 47.68 \\
& Claude-Opus-4.7 & 37.36 & 15.99 & 54.47 & \textbf{56.26} & 55.61 & \textbf{61.26} & 64.14 & 58.37 & 19.20 & 71.20 & 72.15 & 71.53 & 51.74 & 38.75 & 26.47 & 64.29 & 64.50 & \underline{66.77} \\
& GLM-4.6V & 40.95 & 12.07 & 83.67 & 53.61 & \underline{55.74} & 53.59 & 29.26 & 21.10 & 70.75 & 37.89 & 39.80 & 37.94 & 34.67 & 16.92 & 73.29 & 45.17 & 47.18 & 45.19 \\
& Kimi-K2.5 & 22.35 & 7.34 & 36.03 & 41.67 & 37.81 & 52.72 & 47.26 & 44.30 & 32.20 & 55.78 & 54.91 & 59.12 & 35.73 & 27.19 & 33.10 & 49.25 & 46.99 & 56.16 \\
& MiMo-V2.5 & 31.81 & 11.75 & 46.40 & 47.09 & 46.47 & 51.58 & 45.29 & 42.05 & 38.85 & 54.75 & 54.02 & 57.56 & 39.05 & 28.02 & 40.48 & 51.20 & 50.53 & 54.79 \\
& TongUI-7B & 1.96 & 0.49 & \underline{20.00} & 1.96 & 1.96 & 1.96 & 11.81 & 11.81 & 13.40 & 11.81 & 11.81 & 11.81 & 7.25 & 6.57 & 14.29 & 7.25 & 7.25 & 7.25 \\
& UI-TARS-7B-DPO+Proactive & 35.89 & 9.46 & 64.91 & 43.47 & 45.35 & 42.74 & 14.21 & 11.11 & 64.24 & 16.95 & 18.00 & 16.46 & 24.25 & 10.35 & 64.52 & 29.23 & 30.67 & 28.63 \\
& Qwen3.5-9B & 8.81 & 2.94 & 70.73 & 13.82 & 13.72 & 14.76 & 9.28 & 7.17 & 75.42 & 11.90 & 12.34 & 11.98 & 9.06 & 5.21 & 74.55 & 12.79 & 12.98 & 13.27 \\
& ProactiveMobile (7B) & 34.75 & 17.62 & 23.78 & 43.39 & 44.13 & 44.02 & 69.06 & 32.49 & \underline{9.07} & 72.34 & 72.60 & 72.68 & 53.17 & 25.60 & \underline{12.61} & 58.93 & 59.42 & 59.41 \\
& Qwen3.5-9B+Proactive & 36.54 & 13.87 & 25.15 & 40.62 & 41.60 & 40.24 & \underline{70.04} & \underline{64.98} & 11.11 & \underline{73.64} & \underline{74.97} & \underline{73.08} & \underline{54.53} & \underline{41.31} & 14.67 & 58.35 & 59.52 & 57.87 \\
& \textbf{PRPF (Ours)} & \textbf{43.23} & 17.46 & \textbf{19.64} & 50.24 & 51.33 & 50.27 & \textbf{81.72} & \textbf{78.90} & \textbf{1.10} & \textbf{86.69} & \textbf{87.20} & \textbf{87.04} & \textbf{63.90} & \textbf{50.45} & \textbf{5.48} & \textbf{69.81} & \textbf{70.59} & \textbf{70.02} \\
\midrule

\multirow{13}{*}{\textbf{L3}} 
& GPT-5.5 & 37.88 & \underline{17.35} & 29.14 & \underline{51.06} & \textbf{52.98} & \underline{51.54} & 38.93 & 31.47 & 26.45 & 53.10 & 55.58 & 53.08 & 38.34 & 23.53 & 27.86 & \underline{51.95} & \underline{54.12} & \underline{52.22} \\
& o3 & 37.06 & 15.62 & 33.00 & 49.74 & \underline{52.20} & 49.64 & 36.83 & 29.49 & 29.70 & 52.03 & 55.42 & 51.19 & 36.96 & 21.70 & 31.46 & 50.74 & 53.61 & 50.32 \\
& Gemini-3.1-Pro & 15.90 & 8.27 & 72.96 & 33.55 & 32.88 & 37.78 & 25.99 & 16.78 & 64.80 & 47.32 & 48.46 & 50.53 & 20.32 & 12.00 & 68.45 & 39.58 & 39.70 & 43.37 \\
& Claude-Opus-4.7 & 35.97 & 14.90 & 33.13 & \textbf{51.82} & 51.35 & \textbf{55.27} & 39.74 & 30.54 & 27.10 & \underline{57.32} & \underline{59.55} & \underline{58.50} & 37.62 & 21.75 & 30.41 & \textbf{54.23} & \textbf{54.94} & \textbf{56.68} \\
& GLM-4.6V & 32.06 & 10.26 & 69.14 & 45.47 & 48.62 & 44.85 & 26.46 & 14.10 & 68.47 & 43.86 & 47.27 & 43.77 & 29.61 & 11.95 & 68.83 & 44.77 & 48.03 & 44.38 \\
& Kimi-K2.5 & 26.25 & 9.99 & 26.35 & 41.00 & 38.44 & 48.93 & 24.48 & 19.46 & 40.43 & 44.21 & 41.42 & 52.20 & 25.47 & 14.14 & 32.36 & 42.40 & 39.74 & 50.36 \\
& MiMo-V2.5 & 30.25 & 11.35 & 35.87 & 42.24 & 42.83 & 43.90 & 27.77 & 21.35 & 37.66 & 45.86 & 45.34 & 50.14 & 29.16 & 15.73 & 36.64 & 43.83 & 43.93 & 46.63 \\
& TongUI-7B & 1.64 & 0.27 & \textbf{14.29} & 1.64 & 1.64 & 1.64 & 5.13 & 4.90 & 20.00 & 5.14 & 5.13 & 5.25 & 3.17 & 2.30 & 18.42 & 3.17 & 3.17 & 3.22 \\
& UI-TARS-7B-DPO+Proactive & 27.52 & 8.17 & 61.45 & 35.92 & 39.03 & 34.60 & 12.01 & 6.18 & 78.21 & 16.26 & 17.41 & 15.81 & 20.73 & 7.30 & 65.29 & 27.31 & 29.56 & 26.37 \\
& Qwen3.5-9B & 7.36 & 2.73 & 73.45 & 12.41 & 13.05 & 12.66 & 8.04 & 4.78 & 72.73 & 12.84 & 13.41 & 13.21 & 7.66 & 3.62 & 73.13 & 12.60 & 13.21 & 12.90 \\
& ProactiveMobile (7B) & 33.24 & 15.08 & 22.63 & 39.79 & 40.77 & 40.21 & 39.04 & 17.37 & \underline{8.75} & 44.50 & 45.28 & 44.85 & 35.78 & 16.08 & 16.08 & 41.85 & 42.74 & 42.24 \\
& Qwen3.5-9B+Proactive & \textbf{41.24} & 16.44 & \underline{15.26} & 45.34 & 47.18 & 44.48 & \underline{42.54} & \underline{36.48} & 14.74 & 50.91 & 53.67 & 49.65 & \underline{41.81} & \underline{25.22} & \underline{15.03} & 47.78 & 50.03 & 46.75 \\
& \textbf{PRPF (Ours)} & \underline{39.24} & \textbf{17.44} & 16.10 & 44.87 & 46.13 & 44.69 & \textbf{51.87} & \textbf{46.15} & \textbf{3.20} & \textbf{60.47} & \textbf{62.26} & \textbf{60.51} & \textbf{44.77} & \textbf{30.02} & \textbf{9.68} & 51.70 & 53.19 & 51.62 \\
\midrule

\multirow{13}{*}{\textbf{Avg}} 
& GPT-5.5 & \underline{40.34} & \textbf{18.40} & 33.48 & \underline{53.37} & \textbf{54.59} & \underline{54.59} & 55.09 & 49.02 & 19.24 & 65.30 & 66.90 & 65.44 & 47.71 & 33.69 & 23.84 & 59.33 & 60.74 & 60.01 \\
& o3 & 39.63 & \underline{17.25} & 41.45 & 51.80 & \underline{53.89} & 51.85 & 51.31 & 44.53 & 26.89 & 61.05 & 63.27 & 60.50 & 45.46 & 30.87 & 31.61 & 56.42 & 58.58 & 56.17 \\
& Gemini-3.1-Pro & 18.94 & 9.33 & 76.47 & 36.51 & 35.45 & 41.20 & 33.97 & 26.75 & 56.60 & 49.51 & 50.13 & 51.99 & 26.45 & 18.03 & 61.11 & 43.00 & 42.78 & 46.59 \\
& Claude-Opus-4.7 & 37.23 & 15.83 & 41.03 & \textbf{54.03} & 53.53 & \textbf{57.91} & 55.91 & 48.69 & 18.81 & \underline{67.08} & \underline{68.57} & \underline{67.74} & 46.56 & 32.24 & 26.11 & \underline{60.55} & \underline{61.04} & \textbf{62.82} \\
& GLM-4.6V & 36.25 & 11.74 & 74.45 & 49.33 & 52.02 & 49.00 & 29.98 & 20.19 & 67.04 & 41.79 & 44.20 & 41.77 & 33.12 & 15.96 & 69.33 & 45.56 & 48.11 & 45.39 \\
& Kimi-K2.5 & 25.33 & 9.33 & 30.02 & 41.78 & 38.78 & 50.88 & 39.50 & 35.56 & 32.51 & 52.39 & 50.70 & 57.59 & 32.40 & 22.43 & 31.64 & 47.08 & 44.73 & 54.23 \\
& MiMo-V2.5 & 32.10 & 11.63 & 39.45 & 45.17 & 45.28 & 47.85 & 39.08 & 34.54 & 37.06 & 51.79 & 51.19 & 55.16 & 35.58 & 23.07 & 37.87 & 48.48 & 48.23 & 51.50 \\
& TongUI-7B & 1.75 & 0.33 & \textbf{15.79} & 1.75 & 1.75 & 1.75 & 8.15 & 8.04 & 18.58 & 8.16 & 8.15 & 8.21 & 4.95 & 4.18 & 18.10 & 4.95 & 4.95 & 4.97 \\
& UI-TARS-7B-DPO+Proactive & 30.90 & 8.52 & 63.13 & 38.91 & 41.52 & 37.81 & 14.17 & 9.90 & 64.67 & 17.38 & 18.40 & 16.94 & 22.54 & 9.21 & 63.84 & 28.15 & 29.97 & 27.39 \\
& Qwen3.5-9B & 8.30 & 2.95 & 71.52 & 13.24 & 13.60 & 13.75 & 8.15 & 5.69 & 77.13 & 11.65 & 12.11 & 11.83 & 8.22 & 4.32 & 75.30 & 12.44 & 12.85 & 12.79 \\
& ProactiveMobile (7B) & 33.68 & 15.61 & 23.91 & 40.93 & 41.86 & 41.38 & 56.84 & 26.04 & \underline{8.51} & 60.84 & 61.32 & 61.16 & 45.25 & 20.82 & 13.76 & 50.88 & 51.58 & 51.26 \\
& Qwen3.5-9B+Proactive & 39.52 & 14.85 & 18.54 & 43.55 & 45.06 & 42.87 & \underline{59.52} & \underline{54.32} & 10.63 & 64.96 & 66.77 & 64.15 & \underline{49.51} & \underline{34.56} & \underline{13.49} & 54.24 & 55.90 & 53.50 \\
& \textbf{PRPF (Ours)} & \textbf{40.83} & 17.19 & \underline{17.99} & 47.00 & 48.17 & 46.90 & \textbf{69.20} & \textbf{65.15} & \textbf{1.75} & \textbf{75.32} & \textbf{76.38} & \textbf{75.49} & \textbf{55.00} & \textbf{41.15} & \textbf{7.21} & \textbf{61.14} & \textbf{62.26} & \underline{61.18} \\

\bottomrule
\end{tabular}%
}
\caption{Detailed performance comparison using granular, set-based metrics on the ProactiveMobile test set. We report function-name sequence accuracy (Type-Acc$\uparrow$), success rate (SR$\uparrow$), false trigger rate (FTR$\downarrow$), F1$\uparrow$, Precision (P$\uparrow$), and Recall (R$\uparrow$) across different difficulties and modalities. Best results are in bold, and second-best results are underlined. All scores are reported in percentage (\%).}
\label{tab:appendix_granular_metrics}
\end{table*}

\subsection{Detailed Results Analysis of Three Difficulty Levels}

Following the difficulty division method of the ProactiveMobile dataset, we divide the dataset into three difficulty levels: L1, L2, and L3.
\begin{itemize}[leftmargin=*, labelsep=0.5em, noitemsep, topsep=2pt, parsep=0pt, partopsep=0pt]
    \item \textbf{Level 1 (Easy):} Correctly solved by 4--5 out of 5 reference models.
    \item \textbf{Level 2 (Medium):} Correctly solved by 2--3 out of 5 reference models.
    \item \textbf{Level 3 (Hard):} Correctly solved by 0--1 out of 5 reference models.
\end{itemize}

The results of PRPF and all baseline models are shown in Table~\ref{tab:appendix_granular_metrics}, which also provides more detailed insights:

\paragraph{1. PRPF maintains stable advantages across different difficulty levels.}
As the task difficulty increases from L1 to L3, PRPF achieves ALL SR of 66.31\%, 50.45\%, and 30.02\%, respectively, all outperforming all baselines. In particular, PRPF still maintains a clear lead on L2 and L3, indicating that the gains of the model do not come from overfitting to easy samples, but are also effective in medium- and high-difficulty proactive recommendation scenarios.

\paragraph{2. PRPF can generate higher-quality function sets.}
Averaged across the reported difficulty levels and modalities, PRPF achieves the highest F1 and Precision, while its Recall ranks second among all baselines. This indicates that the improvement of PRPF does not only come from the increase of a small number of exactly matched samples, but also from the overall improvement in the selection quality of function sets, enabling the model to more accurately cover the key functions required by users' latent needs.

\paragraph{3. PRPF demonstrates strong cross-modal applicability.} Across different difficulty levels, PRPF achieves higher ALL SR than the strongest baseline, Qwen3.5-9B+Proactive, in both text and multimodal settings. In the text setting, PRPF improves ALL SR from 54.32\% to 65.15\% compared with Qwen3.5-9B+Proactive, while reducing FTR from 10.63\% to 1.75\%. In the multimodal setting, PRPF also improves SR from 14.85\% to 17.19\%. These results indicate that PRPF provides stable proactive perception and content generation capabilities under both text and multimodal conditions.

\paragraph{4. Multimodal tasks remain a bottleneck.} Although PRPF achieves an SR of 17.19\% on multimodal tasks, outperforming Qwen3.5-9B+Proactive, the absolute SR remains low. Even GPT-5.5, which performs best on multimodal tasks, achieves an SR of only 18.40\%. This further demonstrates the challenge of the ProactiveMobile task and indicates that stronger interface understanding capability is still required in the future to achieve a breakthrough in this domain.

\subsection{Efficiency Benchmark Details}
\label{app:efficiency_protocol}

This appendix details the evaluation protocol for the efficiency analysis in Section~\ref{sec:efficiency_analysis}. All reported values are computed on the ProactiveMobile test split and aggregated per test sample after applying the routing policy of each configuration.

\paragraph{Compute Estimation.}

For each PAR invocation, we estimate FLOPs from the actual prompt and generation lengths observed during inference. Let $T_p$ denote the prompt length, $T_g$ the generated length, $N_p$ the number of model parameters, and $L$ and $d$ the number of transformer layers and hidden size. The PAR FLOPs are estimated as
\begin{equation}
F_{\mathrm{PAR}}
=
2N_p(T_p+T_g)
+
4LdT_p(T_p+T_g),
\end{equation}
where the first term approximates parameter-dominated dense computation, while the second term accounts for the attention cost induced by long multimodal prompts.

MPP is measured as an end-to-end front-end, including text encoding, image encoding, and the slow--fast multimodal fusion network. Its measured per-sample cost is 23.58 GFLOPs, 14.16 ms latency, and 1.51 GB peak memory. Let $r$ denote the empirical fraction of test samples routed from MPP to PAR:
\begin{equation}
r
=
\frac{N_{\mathrm{trig}=1}}{N}
=
\frac{2{,}340}{3{,}660}
=
0.639.
\end{equation}

\paragraph{Aggregation Across the Test Set.}

Compute and latency are reported as per-sample averages after applying each system's routing policy. The single-stage baseline invokes the proactive VLM on every sample with the full function pool. The w/o Compression variant keeps the MPP intervention gate but passes the full function pool to PAR, so PAR is invoked only on the $r$ fraction accepted by the gate. The w/o Recommend variant disables the intervention shortcut but keeps MPP's candidate-function filtering, so PAR is invoked on every sample with the candidate function pool. Full PRPF combines both mechanisms: the intervention gate skips no-intervention observations, and candidate-function filtering reduces the function pool for accepted contexts.

\begin{table}[t]
\centering
\footnotesize
\setlength{\tabcolsep}{3pt}
\renewcommand{\arraystretch}{1.15}
\begin{tabular}{@{}lcc@{}}
\toprule
\textbf{Method} & \textbf{Compute} & \textbf{Latency} \\
\midrule

ProactiveMobile (7B)    & $F_{\mathrm{PAR}}^{\mathrm{full}}$                                 & $T_{\mathrm{PAR}}^{\mathrm{full}}$ \\
Full PRPF               & $F_{\mathrm{MPP}}{+}rF_{\mathrm{PAR}}^{\mathrm{comp}}$             & $T_{\mathrm{MPP}}{+}rT_{\mathrm{PAR}}^{\mathrm{comp}}$ \\
w/o Compression         & $F_{\mathrm{MPP}}{+}rF_{\mathrm{PAR}}^{\mathrm{full}}$             & $T_{\mathrm{MPP}}{+}rT_{\mathrm{PAR}}^{\mathrm{full}}$ \\
w/o Recommend           & $F_{\mathrm{MPP}}{+}F_{\mathrm{PAR}}^{\mathrm{comp}}$              & $T_{\mathrm{MPP}}{+}T_{\mathrm{PAR}}^{\mathrm{comp}}$ \\
\bottomrule
\end{tabular}
\caption{Aggregation rules for per-sample compute and latency in the efficiency benchmark.}
\label{tab:efficiency_aggregation}
\end{table}

\paragraph{Memory and Latency Measurement.}

Peak memory is reported as the maximum active stage because MPP and PAR are invoked sequentially: for the single-stage baseline ProactiveMobile (7B),
$M = M_{\mathrm{PAR}}$;
for the PRPF variants,
$M = \max(M_{\mathrm{MPP}}, M_{\mathrm{PAR}})$.
This convention reflects the deployed cascade in which the lightweight pre-reasoning stage and the heavy PAR stage are not treated as one simultaneous forward pass. A conservative co-resident upper bound can be obtained by summing the two stage memories, but the main efficiency table reports the sequential peak.

Latency is measured with the same batched inference path used by the evaluation pipeline. For each configuration, PAR latency is the total wall-clock inference time divided by the number of samples that enter PAR;  for observations rejected by the intervention gate, the latency aggregation includes the measured MPP cost and no PAR forward pass.

\section{Failure Case Analysis}
\label{app:failure}

This appendix provides the failure analysis referenced in
Section~\ref{sec:case_study} and complements the multimodal
limitation discussed in Limitations. All numbers reported here are
computed on the ProactiveMobile test set ($N=3{,}660$) and correspond to
the PRPF row of Table~\ref{tab:overall_performance}: ALL Type-Acc
$=55.00$, SR $=41.15$, FTR $=7.21$. The analysis is
conducted on the same predictions, so the aggregate numbers in this
section can be reconciled with the main table by composition.

\subsection{Error Categorisation}
\label{app:failure_categories}

We use the trigger gate output $g\in\{0,1\}$ from MPP, the predicted function-name sequence $\hat{S}$ from PAR, and the best-matched gold sequence $S^\star$ defined in Appendix~\ref{app:grpo_reward}. A prediction is counted as a Type-Acc error when $\hat{S}$ is not order- and length-equal to $S^\star$. Errors are partitioned into four mutually exclusive categories:
\begin{itemize}[leftmargin=*]
\item \textbf{Non-empty mismatch}: $|\hat{S}|>0$ and $\hat{S}\neq S^\star$.
\item \textbf{Post-gate abstention}: $g=1$ and $|\hat{S}|=0$ while
$|S^\star|>0$. PAR was invoked but the policy declined to recommend.
\item \textbf{Gate misfire}: $g=0$ while the best-matched gold intent
requires recommendation. PAR is never invoked, so no recovery is
possible.
\item \textbf{Parse failure}: $\hat{S}$ cannot be parsed back into a
function-call sequence.
\end{itemize}

\begin{table}[t]
\centering
\small
\setlength{\tabcolsep}{4pt}
\begin{tabular}{@{}lrr@{}}
\toprule
\textbf{Category} & \textbf{Count} & \textbf{\% of Errors} \\
\midrule
Non-empty mismatch        & 751 & 45.6 \\
Post-gate abstention      & 689 & 41.8 \\
Gate misfire              & 200 & 12.1 \\
Parse failure             &   7 & 0.4 \\
\midrule
Total Type-Acc errors     & 1{,}647 & 100.0 \\
\bottomrule
\end{tabular}
\caption{Error categorisation for PRPF on the ProactiveMobile test set.
Parse failures are reported as orphan items because they fall outside
the four well-formed prediction quadrants.}
\label{tab:appendix_error_buckets}
\end{table}

The non-empty mismatch bucket can be further decomposed by comparing
the predicted set against the best-matched gold set
(Table~\ref{tab:appendix_mismatch_subpatterns}). \emph{Off-scenario}
mismatch dominates (40.6\% of non-empty mismatches): PAR returns a
syntactically valid call sequence whose function names share no overlap
with $S^\star$, indicating that the failure is at the intent-selection
level rather than the argument-filling level.

\begin{table}[t]
\centering
\small
\setlength{\tabcolsep}{4pt}
\begin{tabular}{@{}lrr@{}}
\toprule
\textbf{Sub-pattern}                    & \textbf{Count} & \textbf{\%} \\
\midrule
Off-scenario ($\hat{S}\cap S^\star=\emptyset$)        & 305 & 40.6 \\
Spurious ($S^\star=\emptyset,\hat{S}\neq\emptyset$) & 124 & 16.5 \\
Missed steps ($\hat{S}\subsetneq S^\star$)          & 154 & 20.5 \\
Extra steps ($S^\star\subsetneq\hat{S}$)            &  80 & 10.7 \\
Reordering (same set, wrong order)                  &  56 &  7.5 \\
Partial overlap                                     &  32 &  4.3 \\
\midrule
Total                                                & 751 & 100.0 \\
\bottomrule
\end{tabular}
\caption{Sub-patterns of non-empty mismatch errors. The dominant
failure mode is off-scenario misrouting, not argument-level error.}
\label{tab:appendix_mismatch_subpatterns}
\end{table}

\subsection{Where Errors Concentrate: Modality and Gating Path}
\label{app:failure_modality_gate}

To localize the bottleneck, we cross-tabulate Type-Acc by modality
(TEXT vs Multimodal) and gate decision ($g{=}0$ silenced vs $g{=}1$ routed to
PAR), shown in Table~\ref{tab:appendix_modality_gate}.

\begin{table}[t]
\centering
\footnotesize
\setlength{\tabcolsep}{4pt}
\begin{tabular}{@{}lcccc@{}}
\toprule
\multirow{2}{*}{\textbf{Modality}}
  & \multicolumn{2}{c}{$g{=}0$ \textbf{(Silenced)}}
  & \multicolumn{2}{c}{$g{=}1$ \textbf{(Routed To PAR)}} \\
\cmidrule(lr){2-3} \cmidrule(lr){4-5}
& $n$ & Type-Acc & $n$ & Type-Acc \\
\midrule
TEXT       & 1{,}235 & 0.866 & 593     & 0.329 \\
Multimodal & 85      & 0.588 & 1{,}747 & 0.400 \\
\bottomrule
\end{tabular}
\caption{Type-Acc cross-tabulated by modality and gating path.
Most TEXT samples ($1{,}235/1{,}828=67.6\%$) are silenced by the gate
at high accuracy. The bottleneck shifts to PAR once a sample is routed:
$g{=}1$ accuracy is below $0.41$ for both modalities.}
\label{tab:appendix_modality_gate}
\end{table}

Two observations matter for the multimodal-bottleneck argument. First,
the $g{=}0$ silenced path is concentrated on TEXT ($93.6\%$ of silenced
samples are TEXT), where the trigger decision is comparatively easy and
acc reaches $0.866$. The high overall TEXT accuracy reported in the
main table is largely attributable to this silence dividend rather than
to stronger TEXT recommendation by PAR. Second, on the $g{=}1$ path
TEXT accuracy ($0.329$) is in fact \emph{lower} than Multimodal accuracy
($0.400$): the easy TEXT cases have already been gated out, and what
remains for PAR is the harder TEXT residual. This indicates that
further gains require improving PAR's reasoning on the routed subset,
not pushing the gate to filter more aggressively.

\subsection{Intent-Scenario-Level Concentration}
\label{app:failure_scene}

The 14 high-level intent scenarios used by MPP differ markedly in error
profile. We group them into three substantive clusters
(Table~\ref{tab:appendix_scene_clusters}); two scenes with very small
$n$ (Logistics \& Delivery, $n=54$; Smart Home, $n=10$) are excluded
from clustering to avoid noise.

\begin{table*}[!htbp]
\centering
\small
\setlength{\tabcolsep}{6pt}
\renewcommand{\arraystretch}{1.25}

\begin{tabularx}{\textwidth}{
@{}
p{0.24\textwidth}
>{\raggedright\arraybackslash}X
>{\centering\arraybackslash}p{0.08\textwidth}
>{\centering\arraybackslash}p{0.10\textwidth}
>{\centering\arraybackslash}p{0.08\textwidth}
@{}
}
\toprule
\textbf{Cluster}
& \textbf{Representative Intent Scenario}
& \textbf{$n$}
& \textbf{Type-Acc}
& $\bar{\Delta}_{T-M}$ \\
\midrule

I. Multimodal-hard
& Sports \& Health, Entertainment \& Media, Office Work, Social Communication
& 1{,}200
& 0.645
& $+0.41$ \\

\addlinespace[6pt]

II. Domain-knowledge-hard
& Travel \& Lodging, Food \& Dining, Transportation
& 997
& 0.431
& $+0.18$ \\

\addlinespace[6pt]

III. Balanced
& Device \& System Management, Shopping, Personal Management, Content Creation, Financial Services
& 1{,}399
& 0.557
& $+0.24$ \\

\bottomrule
\end{tabularx}

\caption{Intent-scenario-level clusters with mean Type-Acc and mean TEXT$-$Multimodal gap $\bar{\Delta}_{T-M}$. Cluster~II is the largest absolute drag on overall accuracy and shows a small modality gap, indicating that even the TEXT branch underperforms in these scenes. Cluster~I has high overall accuracy but the largest modality gap, so further gains specifically require stronger GUI grounding.}
\label{tab:appendix_scene_clusters}
\end{table*}

The off-scenario mismatch sub-pattern is highly concentrated in
Travel \& Lodging (60), Shopping (47), and Personal
Management (44) together account for $151/305=49.5\%$ of all off-scenario
errors, suggesting that the residual errors in these scenarios are
disproportionately rooted in confusable functions across closely
related lifestyle scenes (e.g.\ booking, ordering, ride-hailing) rather
than in interface understanding.

\subsection{A Reference Multimodal Success Case}
\label{app:failure_cases}

To complement the failure statistics above with a concrete example of successful behavior, Table~\ref{tab:case_c} reproduces a single multimodal sample
on which PRPF achieves SR$=1$ in a Financial Services intent scenario---the
behavior the system already exhibits when the gate routes correctly
and PAR's \texttt{<thinking>} and \texttt{<function\_selection>} both
align with the gold function. The table reproduces the abridged
benchmark inputs---user profile~$U$, device state~$D$, world
context~$W$, and trace~$H$---together with PAR's parsed structured
output as
\texttt{<ui\_summary>} $\rightarrow$ \texttt{<thinking>} $\rightarrow$ \texttt{<recommendations>}.
The \texttt{<thinking>} block reproduces both the free-form reasoning
and the embedded \texttt{<function\_selection>} sub-block. The case is
rendered as a single-column \texttt{longtable} in Table~\ref{tab:case_c}. The original ProactiveMobile data is in Chinese;
we render all natural-language content in English here for
readability, with function names, parameter keys, and structural tags
preserved verbatim.

\subsection{Implications}
\label{app:failure_implications}

The category- and slice-level numbers above isolate two complementary
residual problems for PRPF. (i)~On the modality axis, $g{=}1$ Multimodal
samples remain the dominant absolute error mass and the multimodal
limitation reported in the Limitations section is quantitatively
concentrated in Cluster~I, where TEXT accuracy already exceeds $0.80$.
Closing this gap will mainly require stronger GUI grounding inside PAR.
(ii)~On the intent-selection axis, the off-scenario sub-patten indicates that PAR's failures in
travel-/shopping-/management-style intent scenarios are routing failures rather
than argument-level failures. This points to expanding the function set
seen during training and refining the candidate-pool compression in
MPP, rather than further reward shaping at the GRPO stage.
\section{Prompts for LLM Agents}
\label{app:prompts}

At inference time, the trained PAR model (Section~\ref{sec:par})
consumes a single user-message prompt template assembled by the
eval-data builder: each ProactiveMobile sample is rewritten into one
turn that bundles the per-sample context $(U, D, W)$, the
MPP-predicted intent scenarios, and the MPP-restricted function pool,
and the resulting message is wrapped with the standard Qwen-3 chat
template before being fed to vLLM. Only samples for which the MPP
trigger gate predicts $g=1$ are routed through this prompt;
gate-silenced samples are short-circuited to the empty-recommendation
output without a forward pass through PAR. The sampler-side
conversation is built with a fixed system message
\texttt{"You are a helpful assistant."} followed by a single user
message that concatenates a Chinese task header, an XML-tagged
context bundle, the MPP-predicted scenes, the available function
pool, and a structured-output instruction tail; for multimodal
samples the leading \texttt{<image>} placeholders are stripped from
the textual content and the corresponding screenshot files are
attached as \texttt{image\_url} entries so that vLLM places the
image tokens at the same position relative to the textual context
as during SFT. A sample at the
end of the appendix documents the assembled prompt for the multimodal branch; the original prompts are in Chinese, and we
render them in English here for readability while preserving JSON
keys, structural tags, and function names verbatim.

\onecolumn
\setlength{\LTcapwidth}{\linewidth}
\setlength{\tabcolsep}{4pt}
\renewcommand{\arraystretch}{1.15}
\begin{longtable}{@{}>{\bfseries\raggedright\arraybackslash\footnotesize}p{2.5cm} >{\raggedright\arraybackslash\footnotesize}p{12.5cm}@{}}
\toprule
\endfirsthead
\toprule
\endhead
User Profile ($U$) &
A 35-year-old technology-industry professional with seven years of
investing experience; holds shares in Zoom, Apple, Tesla, Amazon, and
other tech companies... \\
\midrule
Device Status ($D$) &
Time 12:48; WiFi connected; battery sufficient. Multiple finance apps
installed: Yahoo Finance, Investing.com, Bloomberg, TradingView. Over
the past three months the user has averaged 15+ daily opens of
finance apps... \\
\midrule
World Information ($W$) &
Currently in market hours; NASDAQ $+2.03\%$, S\&P~500 $+1.02\%$;
market sentiment positive. Tech sector broadly stronger; multiple
cloud-computing and e-commerce stocks up... \\
\midrule
Behavioral Trajectories ($H$) &
\begin{tabular}{@{}ccccc@{}}
\includegraphics[width=0.18\linewidth,height=1.6cm,keepaspectratio]{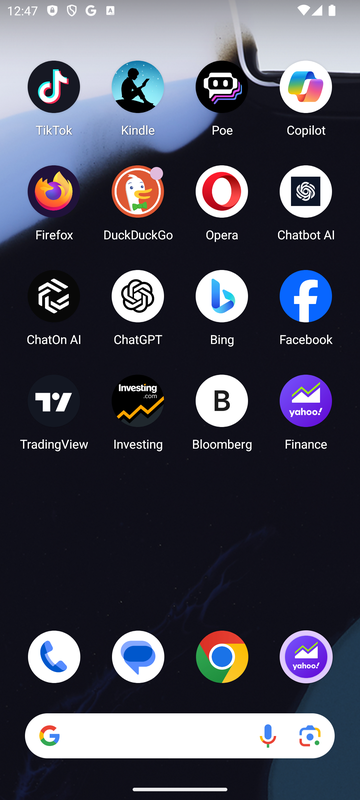} &
\includegraphics[width=0.18\linewidth,height=1.6cm,keepaspectratio]{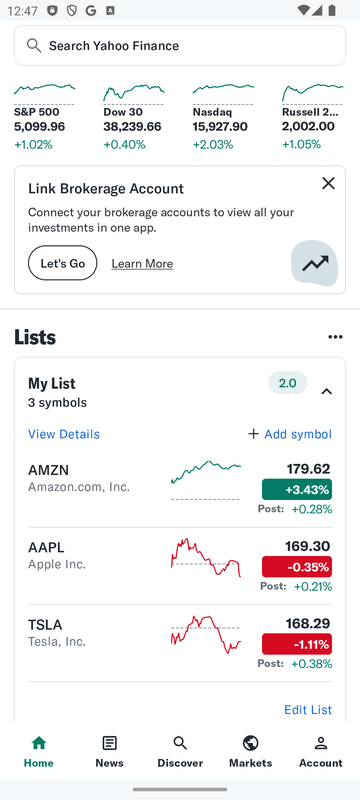} &
\includegraphics[width=0.18\linewidth,height=1.6cm,keepaspectratio]{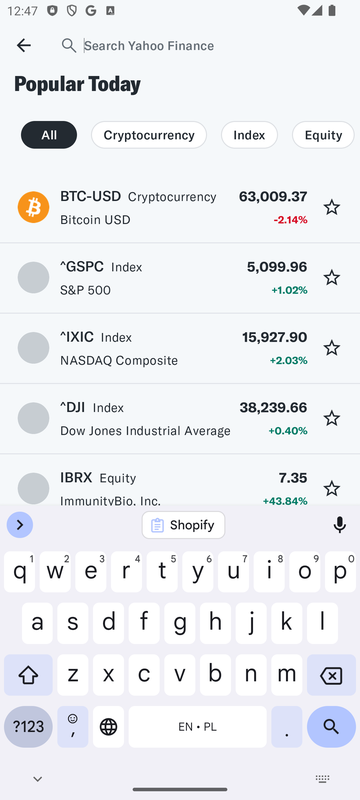} &
\includegraphics[width=0.18\linewidth,height=1.6cm,keepaspectratio]{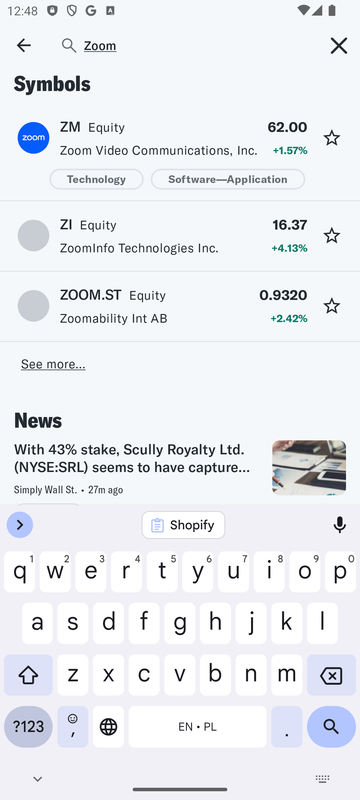} &
\includegraphics[width=0.18\linewidth,height=1.6cm,keepaspectratio]{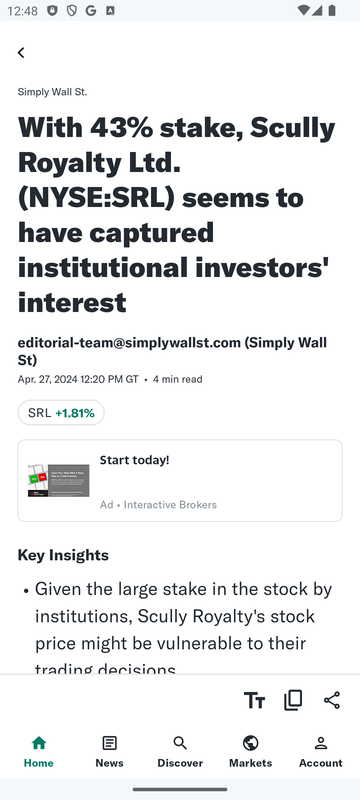} \\
\includegraphics[width=0.18\linewidth,height=1.6cm,keepaspectratio]{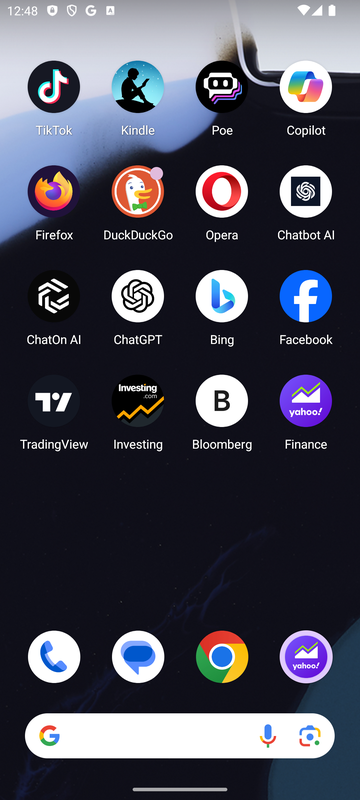} &
\includegraphics[width=0.18\linewidth,height=1.6cm,keepaspectratio]{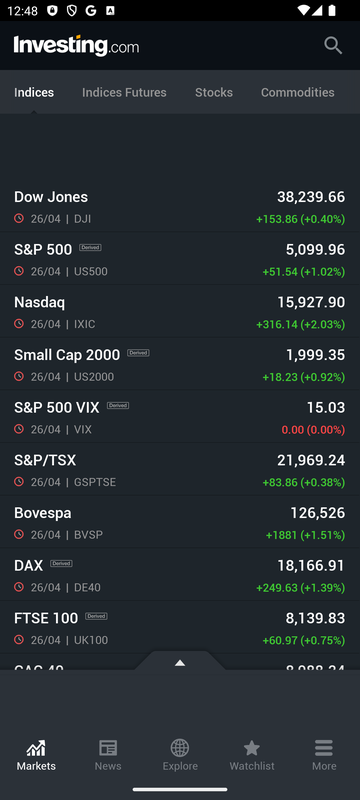} &
\includegraphics[width=0.18\linewidth,height=1.6cm,keepaspectratio]{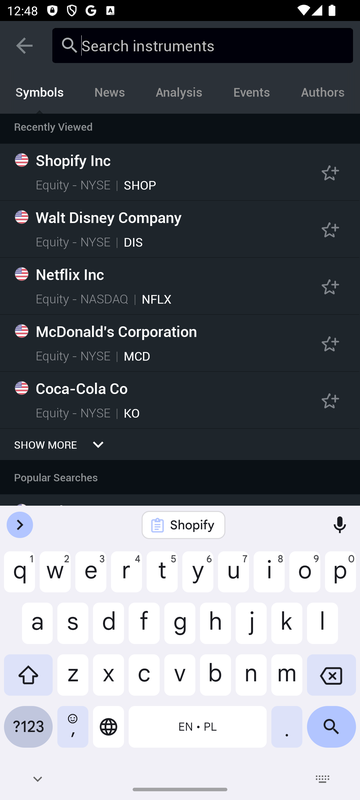} &
\includegraphics[width=0.18\linewidth,height=1.6cm,keepaspectratio]{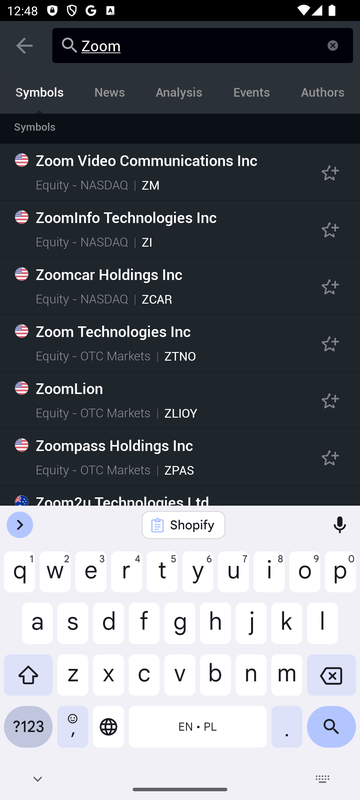} &
\includegraphics[width=0.18\linewidth,height=1.6cm,keepaspectratio]{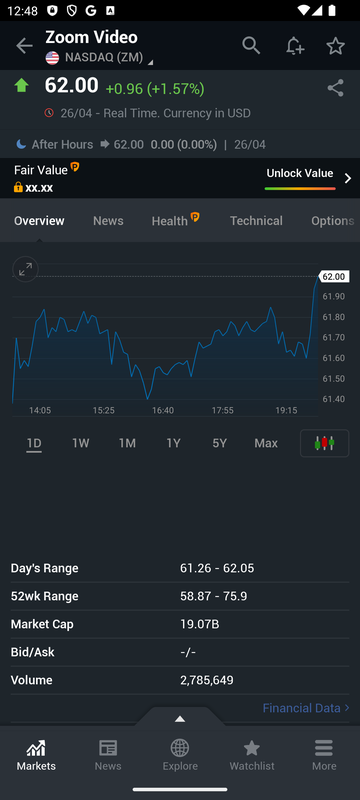}
\end{tabular}\newline
{\scriptsize Frames 1--10: Yahoo Finance $\rightarrow$ search ``Zoom''
$\rightarrow$ Simply Wall St.\ article $\rightarrow$ Investing.com
$\rightarrow$ Zoom Video (NASDAQ:~ZM) ticker page.}
\\
\midrule
\texttt{<ui\_summary>} &
Page progression: from the home screen the user opens Yahoo Finance
and checks ``My List'' (containing AMZN, AAPL, TSLA), enters ``Zoom''
in the search bar, taps a Simply Wall St...
\\
\midrule
\texttt{<thinking>} &
The user is currently viewing the Zoom Video (NASDAQ:~ZM) ticker page
on Investing.com, with price \$62.00 ($+1.57\%$). Earlier he searched
Zoom on Yahoo Finance and read a Simply Wall St.\ article on Scully
Royalty Ltd., then switched platforms. This sequence indicates
multi-platform information gathering at the pre-decision stage...\par

\smallskip\texttt{<function\_selection>}\par
\textbf{Candidate function pool.}
\texttt{search\_financial\_info},
\texttt{create\_financial\_analysis\_view},
\texttt{create\_portfolio},
\texttt{manage\_budget\_and\_plans},
\texttt{bank\_service\_assistance},
...\par
\textbf{Step-by-step reasoning.}
\vspace{-0.6ex}
\begin{itemize}\setlength\itemsep{-0.4ex}\setlength\parsep{0pt}\setlength\topsep{0pt}
\item \texttt{search\_financial\_info}: supports
\texttt{search\_target="analysis\_articles"}, \texttt{search\_keywords="Zoom"},
\texttt{location="US"}; directly matches. \textbf{Match.}
\item \texttt{create\_financial\_analysis\_view}: builds a chart
view, not a search. \textbf{Excluded.}
\item ...
\end{itemize}
\textbf{Selected function and parameters.}
\texttt{search\_financial\_info} with
\texttt{search\_target="analysis\_articles"},
\texttt{search\_keywords="Zoom"},
\texttt{location="US"}.\par
\texttt{</function\_selection>}\par
\texttt{</thinking>}
\\
\midrule
\texttt{<recommen\-dations>} &
\textit{Instruction:} ``Recommend deep analysis articles or research
reports on Zoom.''\par
\smallskip
\textit{Function call (parsed):}\par
{\scriptsize\ttfamily
[\{\newline
\hspace*{1ex}"name":\newline
\hspace*{2ex}"search\_financial\_info",\newline
\hspace*{1ex}"parameters": \{\newline
\hspace*{2ex}"search\_target":\newline
\hspace*{2ex}\hphantom{x}"analysis\_articles",\newline
\hspace*{2ex}"search\_keywords": "Zoom",\newline
\hspace*{2ex}"filter\_criteria": "",\newline
\hspace*{2ex}"sort\_order": "",\newline
\hspace*{2ex}"category": "",\newline
\hspace*{2ex}"include\_analysis": "",\newline
\hspace*{2ex}"location": "US"\newline
\hspace*{1ex}\}\newline
\}]
}
\\
\midrule
Outcome &
Type-Acc match; the LLM judge marks it as passing, \textbf{SR$=1$.}
\\
\bottomrule
\caption{Case study: Multimodal --- Match. Scene: Financial Services. Sample id \texttt{40590079}.}\label{tab:case_c}\\
\end{longtable}

\begin{tcolorbox}[
    title=Prompt $\mathcal{P}_{\text{inf-GUI}}$: PAR Inference Prompt (Multimodal/GUI),
    colback=gray!5!white, colframe=gray!75!black,
    fonttitle=\bfseries,
    label={tab:prompt_inf_gui}]

You are an intelligent assistant. Based on the user profile (profile), device status (phone), environmental information (world), and screenshots, analyze the user’s current behavior and determine whether a recommendation is needed.

\textbf{User profile} 
\\
\textbf{Device status} 
\\
\textbf{Environmental information} 
\\
\textbf{predicted\_scenes}
\{List of predicted scene probabilities, different for each sample\}\\
\textbf{available\_functions}
\{List of available functions and parameter definitions, different for each sample\}
\\
Please output in the following order:\\
\textbf{ui\_summary}: Based on the screenshots, describe the current stage of the user interface, key interaction areas, state changes, and the current stage of the workflow. Include key identifiable entities in the screenshots, such as specific app names, page titles, contacts, file names, etc.\\
\textbf{thinking}: Conduct sufficient step-by-step reasoning covering the following aspects:\\
a) Behavior recognition: Describe what the user is doing based on the screenshots and extract key entities such as app names, file names, contacts, page content, etc.;\\
b) Intent analysis: Combine the user profile (profile), device status (phone), and environmental information (world) to analyze the user’s deeper intent and the current task stage (just started, in progress, completed);\\
c) Need assessment: Determine whether the user currently has any unmet needs or whether there is an optimizable next step. If the task has been completed and there is no obvious follow-up need, clearly conclude that no recommendation is needed;\\
d) If a recommendation is needed, explain what to recommend, why it should be recommended at this moment rather than letting the user complete it themselves, and the key parameter information required for the recommendation, such as target app, operation object, recipient, etc.\\
\textbf{recommendations}: Based on the selected functions and parameters in function\_selection, generate the final function call(s), which may include 0, 1, or multiple functions. The function names and parameter values should be consistent with the reasoning conclusion in function\_selection, in the following format:
[{"instruction": "Recommendation instruction", "function": [{"name": "FunctionName1", "parameters": {...}}, {"name": "FunctionName2", "parameters": {...}}]}]
\\
If there is no recommendation, function\_selection is not needed in thinking; directly output: No recommendation

\end{tcolorbox}

\twocolumn

\end{document}